\documentclass{article}

\usepackage{microtype}
\usepackage{graphicx}
\usepackage{subfigure}
\usepackage{booktabs} % for professional tables
\usepackage{multirow}
\usepackage{makecell}
\usepackage{enumitem}

\usepackage{hyperref}

\usepackage[accepted]{icml2024}

\usepackage{amsmath}
\usepackage{amssymb}
\usepackage{mathtools}
\usepackage{amsthm}
\usepackage{comment}
\usepackage{caption}
\usepackage{subcaption}

\usepackage[capitalize,noabbrev]{cleveref}

\usepackage{listings}

\theoremstyle{plain}

\theoremstyle{definition}

\theoremstyle{remark}

\usepackage[textsize=tiny]{todonotes}

\icmltitlerunning{Rethinking Generative Large Language Model Evaluation for Semantic Comprehension}

\begin{document}

\twocolumn[
\icmltitle{Rethinking Generative Large Language Model Evaluation for\\ Semantic Comprehension}



\icmlsetsymbol{equal}{*}

\begin{icmlauthorlist}
\textbf{Fangyun Wei}\textsuperscript{*}\quad\quad
\textbf{Xi Chen}\textsuperscript{*} \quad\quad
\textbf{Lin Luo}\textsuperscript{*} \\
Microsoft Research Asia \\
\textsuperscript{*}Equal contribution\\
\texttt{\{fawe,xichen6,liluo\}@microsoft.com} \\
Project page: \href{https://luolinrowling.github.io/Rethink-LLM-Eval}{https://luolinrowling.github.io/Rethink-LLM-Eval}\\
\end{icmlauthorlist}

%\icmlkeywords{Machine Learning, ICML}

\vskip 0.3in
]

\begin{abstract}
Despite their sophisticated capabilities, large language models (LLMs) encounter a major hurdle in effective assessment. This paper first revisits the prevalent evaluation method—multiple choice question answering (MCQA), which allows for straightforward accuracy measurement. Through a comprehensive evaluation of 24 models across 11 benchmarks, we highlight several potential drawbacks of MCQA, for instance, the inconsistency between the MCQA evaluation and the generation of open-ended responses in practical scenarios. In response, we introduce an RWQ-Elo rating system, engaging 24 LLMs such as GPT-4, GPT-3.5, Google-Gemini-Pro and LLaMA-1/-2, in a two-player competitive format, with GPT-4 serving as the judge. Each LLM receives an Elo rating thereafter. This system is designed to mirror real-world usage, and for this purpose, we have compiled a new benchmark called ``Real-world questions'' (RWQ), comprising 20,772 authentic user inquiries. Additionally, we thoroughly analyze the characteristics of our system and compare it with prior leaderboards like AlpacaEval and MT-Bench. Our analysis reveals the stability of our RWQ-Elo system, the feasibility of registering new models, and its potential to reshape LLM leaderboards. 
\end{abstract}
\vspace{-5mm}
\section{Introduction}
\vspace{-1mm}
In recent years, the advent of large language models (LLMs)~\cite{gpt1,gpt2,gpt3,touvron2023llama2} has revolutionized the field of artificial intelligence, offering unprecedented capabilities in natural language processing and understanding. However, the rapid development of LLMs also present significant challenges, particularly in terms of evaluation. The effective assessment of these models is crucial to ensure their reliability, fairness, harmlessness, and ethical use.
\vspace{-1mm}

This paper aims to contribute to the ongoing discourse in the field of natural language processing by providing a comprehensive evaluation of LLMs. Prior research has primarily focused on evaluating a range of capabilities including broad world knowledge, commonsense reasoning, and specialized skills like coding and mathematics. This has led to the introduction of various benchmarks (e.g. MMLU~\cite{hendrycks2020measuring}, AGIEval~\cite{zhong2023agieval}, and ARC~\cite{clark2018think}) and assessment platforms (e.g., HELM~\cite{HELM} and Harness~\cite{gao2021framework}). While evaluating specific skills like coding, which can be assessed through targeted coding tasks and test cases using pass rate as a metric~\cite{chen2021evaluating}, and mathematics, where unique solutions enable accuracy-based metrics~\cite{cobbe2021gsm8k}, is relatively clear-cut, assessing open-ended areas such as semantic comprehension, remains challenging. This complexity stems from two primary factors: 1) the open-ended nature of LLM responses leads to a broad spectrum of possible answers; and 2) the non-uniqueness of what constitutes a reasonable answer to a given question.
\vspace{-1mm}

The evaluation of LLMs has predominantly centered around multiple choice question answering (MCQA) due to its straightforward approach in measuring LLM performance via accuracy and its facilitation of comparisons with other LLMs. This paper delves into the inherent shortcomings of the MCQA evaluation. Initially, we highlight the discrepancy between the real-world usage of LLMs for responding to user queries in an open-ended manner, and the confined nature of selecting the best option in MCQA. Moreover, the methodology for generating MCQA predictions differs among models. This can involve either pinpointing the likeliest token (e.g., ``A'' or ``B'') upon reviewing the main question and its associated choices, or calculating the alignment between the question and each option using perplexity. Such variations result in inconsistent evaluations across different LLMs. Additionally, the open-ended responses provided by LLMs may not correspond with their MCQA predictions, leading to unreliable evaluations. This paper comprehensively examines these potential disadvantages through a combination of quantitative and qualitative studies.
\vspace{-1mm}

In practical settings, LLMs perform tasks such as following user instructions, responding to inquiries, and providing answers to questions. We advocate for the use of LLMs to generate open-ended responses and recommend directly assessing these responses. Previous works, such as AlpacaEval~\cite{alpaca_eval}, compare an LLM's response to that of a pre-defined benchmark model to the same query, and then calculate a win-rate, which serves as the comparison metric across various LLMs. Nevertheless, when this benchmark model is significantly superior (e.g. GPT-4-Turbo) or inferior, AlpacaEval may not distinguish performance differences among the LLMs being evaluated (see Figure~\ref{fig:compare_elo_with_alpaca}). Furthermore, the benchmark model serves as an ``intermediary agent'' to compare the relative capabilities between two LLMs—a direct comparison between them is not made. If an optimization objective aimed at surpassing the benchmark model is incorporated into an LLM's training or fine-tuning process, it might result in an inflated win-rate, leading to an artificially high ranking.

In this work, we evaluate the performance of LLMs through a series of two-player contests. Each round involves randomly selecting two LLMs to respond to a query sampled from our ``Real-World Questions'' (RWQ) benchmark, which consists of 20,772 realistic user queries collected from sources such as Google Trends and Quora. A judge then determines the winner and the loser. Each LLM under assessment is assigned a rating that is adjusted after each competition based on the results. We utilize the Elo algorithm to update the ratings, naming this system the RWQ-Elo system. Recently, GPT-4, known for its superior language comprehension abilities, has been incorporated as a judge in numerous tasks~\cite{alpaca_eval}. Our meticulous design of evaluation criteria and guidelines has shown that GPT-4's decisions align with those of human evaluators 95\% of the time. Therefore, GPT-4 is employed as the judge in our RWQ-Elo system to enhance scalability. By orchestrating these contests among 24 LLMs such as GPT-4, GPT-3.5, Google Gemini-Pro and LLaMA family, we could directly evaluate their relative capabilities, making them more distinguishable. Additionally, we analyze the stability of our RWQ-Elo system, its relation to existing LLM leaderboards, and the schema for new model registration. Owing to its simplicity and robustness, we anticipate that our approach could become a new standard for LLM evaluation.
\vspace{-2mm}
\section{Related Work}
\vspace{-1mm}
\textbf{Advancements in Generative Large Language Models.} The advent of generative LLMs marks a new era in the development of sophisticated AI algorithms adept at language comprehension and task execution. The introduction of ChatGPT~\cite{achiam2023gpt} has garnered significant attention, exemplifying this progress. These LLMs explore various architectures, such as causal decoders (for instance, GPT~\cite{gpt1,gpt2,gpt3}, LLaMA~\cite{touvron2023llama,touvron2023llama2}, OPT~\cite{zhang2022opt}, BLOOM~\cite{workshop2022bloom}, LaMDA~\cite{thoppilan2022lamda}), encoder-decoder frameworks (like T5~\cite{raffel2020exploring} and Flan-T5~\cite{chung2022scaling}), and mixture-of-experts models (such as Switch Transformer~\cite{fedus2022switch} and GLaM~\cite{du2022glam}), along with innovative structures (for example, RWKV~\cite{peng2023rwkv} and RetNet~\cite{sun2023retentive}). To reconcile the gap between pre-training objectives and user-directed goals, notably ``follow their instructions helpfully and safely''~\cite{gpt2}, instruction tuning techniques~\cite{ouyang2022training} are introduced. Instruction tuning or supervised fine-tuning in LLaMA, in particular, is key in developing tailored or niche models, including Vicuna~\cite{chiang2023vicuna}, Stanford Alpaca~\cite{taori2023stanford}, WizardLM~\cite{xu2023wizardlm}, and Xwin-LM~\cite{xwin-lm}. Additional research areas encompass scaling LLMs~\cite{hoffmann2022training,gopher,palm}, managing long contexts~\cite{su2024roformer,ding2023longnet}, devising decoding strategies~\cite{leviathan2023fast,chen2023accelerating,li2023rain,li2024eagle}, innovating sampling methods~\cite{fan2018hierarchical,holtzman2019curious}, enhancing training efficiency~\cite{huang2019gpipe,shoeybi2019megatron,rajbhandari2020zero,rasley2020deepspeed,dao2022flashattention,hu2021lora}, foundational operators~\cite{ba2016layer,shazeer2020glu}, and training data collection~\cite{zhu2015aligning,raffel2020exploring,liu2019roberta}, among others.

\textbf{LLM Evaluation.}
Evaluation of LLMs encompasses a variety of domains, including understanding knowledge~\cite{khot2020qasc}, aligning responses to questions and instructions, utilization of tools, safety considerations, and specialized competencies in areas such as programming~\cite{chen2021evaluating}, mathematics~\cite{cobbe2021gsm8k,austin2021program}, and language translation~\cite{bojar2014findings,bojar2016findings,lison2016opensubtitles2016}. A fundamental skill for LLMs is the ability to possess extensive general knowledge and to respond to questions or queries both correctly and logically. To evaluate this skill, numerous benchmarks have been developed, including human examination datasets (e.g., e.g. MMLU~\cite{hendrycks2020measuring}, AGIEval~\cite{zhong2023agieval}, C-Eval~\cite{huang2023ceval}, and RACE~\cite{lai2017race}), datasets for assessing commonsense reasoning (e.g., ARC~\cite{clark2018think} and CommonSenseQA~\cite{talmor2018commonsenseqa}), and question-answering datasets (e.g., PIQA~\cite{bisk2020piqa}, OpenBookQA~\cite{mihaylov2018can} and BoolQ~\cite{clark2019boolq}). Among these benchmarks, multiple choice question answering (MCQA) is particularly prominent due to the straightforward nature of using accuracy as a measurement criterion. In order to consolidate different benchmarks and offer a unified interface, various evaluation platforms have been introduced. These include HELM~\cite{HELM}, Harness~\cite{gao2021framework}, OpenCompass~\cite{2023opencompass} and Big-bench~\cite{srivastava2023beyond}. Diverging from MCQA, AlpacaEval~\cite{alpaca_eval} proposes a different approach: having LLMs compete against stronger counterparts, namely Text-Davinci-003 and GPT-4, in a question-answering task, with GPT-4 serving as an adjudicator owing to its advanced capabilities. The evaluation metric adopted is the win-rate of these LLMs against Text-Davinci-003 or GPT-4. In contrast to AlpacaEval, which advocates for an LLM to compete with a single model, this work presents a multiple-player Elo rating system.
\vspace{-2mm}
\section{Rethinking MCQA Evaluation}
\vspace{-1mm}
\label{sec:MCQAEval}

\begin{table*}[!t]
    \centering
    \caption{Averaged results across 11 0-shot datasets including MMLU~\cite{hendrycks2020measuring}, HellaSwag~\cite{zellers2019hellaswag}, ARC-Challenge and -Easy~\cite{clark2018think}, BoolQ~\cite{clark2019boolq}, SIQA~\cite{sap2019socialiqa}, PIQA~\cite{bisk2020piqa}, AGIEval (English only)~\cite{zhong2023agieval}, OpenBookQA (with fact)~\cite{mihaylov2018can}, CommonSenseQA~\cite{talmor2018commonsenseqa} and RACE (all)~\cite{lai2017race}, using 7 evaluation strategies introduced in Section~\ref{sec:MCQAEval}. We use the latest models up to February 1, 2024. We use general MCQA prompt for all benchmarks without dedicated design. Detailed results for each benchmark can be found in Table~\ref{tab:MMLU-0-shot}-\ref{tab:RACE-all-0-shot}.}
    \vspace{-3mm}
        \begin{tabular}{ccccccccc}
            \toprule
            Model & Size & Choices & \makecell{Choices\\(Circular)} & Vocab & \makecell{Vocab\\(Circular)} & Alignment & \makecell{Normalized \\Alignment} & PPL \\
             \midrule
             \multirow{2}{*}{\makecell{MPT\\~\cite{team2023introducing, mosaicml2023introducing}}} & 7B & 36.0 & 2.2 & 35.2 & 2.0 & 52.3 & 54.3 & 54.2 \\
                & 30B & 53.0 & 26.4 & 49.2 & 23.0 & 54.8 & 57.1 & 56.8\\
             \midrule
            \makecell{MPT-Chat\\~\cite{mosaicml2023introducing}} & 30B & 61.5 & 37.9 & 60.8 & 37.0 & 56.7 & 58.9 & 58.3 \\
            \midrule
             \multirow{2}{*}{\makecell{Falcon\\~\cite{almazrouei2023falcon}}} & 7B & 31.7 & 3.6 & 30.2 & 2.9 & 52.3 & 54.2 & 54.7 \\
                & 40B & 62.3 & 36.6 & 62.0 & 36.4 & 57.8 & 59.4 & 59.8 \\
             \midrule
             \multirow{4}{*}{\makecell{LLaMA-1\\~\cite{touvron2023llama}}} & 7B & 40.4 & 8.0 & 38.7 & 7.2 & 52.8 & 54.7 & 53.6 \\
                & 13B & 52.6 & 20.1 & 50.2 & 18.3 & 54.6 & 56.1 & 55.3 \\
                & 30B & 65.6 & 45.3 & 65.4 & 45.0 & 57.0 & 58.7 & 57.8 \\
                & 65B & 67.5 & 45.2 & 66.1 & 43.9 & 58.3 & 60.1 & 59.4 \\
            \midrule
            \multirow{3}{*}{\makecell{LLaMA-2\\~\cite{touvron2023llama2}}} & 7B & 47.5 & 17.4 & 42.7 & 14.1 & 53.3 & 55.1 & 54.4 \\
                    & 13B & 60.8 & 31.1 & 58.6 & 29.7 & 55.5 & 57.0 & 56.4 \\
                    & 70B & 75.2 & 58.4 & 74.8 & 57.9 & 59.0 & 60.4 & 59.8 \\
            \midrule
            \multirow{3}{*}{\makecell{LLaMA-2-Chat\\~\cite{touvron2023llama2}}} & 7B & 57.7 & 28.8 & 55.8 & 28.3 & 54.1 & 55.8 & 54.5 \\
                         & 13B & 65.4 & 40.9 & 65.3 & 40.8 & 56.0 & 58.6 & 57.0 \\
                         & 70B & 74.3 & 56.8 & 74.2 & 56.6 & 58.9 & 60.7 & 59.5 \\
            \midrule
            \multirow{2}{*}{\makecell{WizardLM\\~\cite{xu2023wizardlm}}} 
                   & 13B & 67.6 & 47.1 & 67.6 & 47.1 & 56.6 & 58.1 & 57.4 \\
                   & 70B & 76.7 & 61.7 & 76.6 & 61.6 & 59.2 & 60.3 & 59.8 \\
            \midrule
            \multirow{2}{*}{\makecell{Xwin-LM\\~\cite{xwin-lm}}} & 7B & 55.0 & 25.2 & 54.8 & 25.0 & 55.0 & 55.9 & 55.3 \\
                 & 13B & 64.0 & 34.9 & 63.9 & 34.7 & 57.3 & 58.6 & 58.1 \\
            \midrule
            \multirow{2}{*}{\makecell{Alpaca\\~\cite{taori2023alpaca}}} & 7B & 52.7 & 24.4 & 52.5 & 24.1 & 54.4 & 56.5 & 55.1 \\
                   & 13B & 54.0 & 30.3 & 53.6 & 30.0 & 55.5 & 56.8 & 55.9 \\
            \midrule
            \multirow{3}{*}{\makecell{Vicuna\\~\cite{chiang2023vicuna}}} & 7B & 62.6 & 41.1 & 62.5 & 41.0 & 53.8 & 54.7 & 54.3 \\
                   & 13B & 68.8 & 50.1 & 68.7 & 50.1 & 56.3 & 57.6 & 56.6 \\
                   & 33B & 69.6 & 50.2 & 64.3 & 45.0 & 56.4 & 57.9 & 57.4 \\
            \bottomrule        
        \end{tabular}
        \vspace{-3mm}
    \label{tab:MCQA_avg}
\end{table*}

\textbf{Formulation and Notations.} Multiple choice question answering (MCQA) has emerged as the dominant evaluation task, favored for its convenience in quantitative assessment. Generally, a multiple-choice question consists of a question $\mathcal{Q}$, $K$ choices $\{\mathcal{C}_i\}_{i=1}^{K}$, and a reference answer choice $\mathcal{A}$. Each choice $\mathcal{C}_i$ (e.g. ``A. 0 degrees Celsius.'') comprises a choice number (e.g. ``A'') and a statement $\mathcal{S}_i$ (e.g. ``0 degrees Celsius.'').  An LLM is supposed to predict the most accurate answer choice upon encountering $\mathcal{Q}$ and $\{\mathcal{C}_i\}_{i=1}^{K}$. If the prediction matches $\mathcal{A}$, the corresponding question is considered correctly answered. Consequently, accuracy is readily adopted as the evaluation metric. Additionally, we use $\mathcal{V}$ to denote the entire vocabulary of an LLM, and $\mathcal{V}_{\mathcal{C}}$ to represent the set containing all choice number tokens (e.g., \{``A'', ``B'', ``C'', ``D''\} for MMLU and \{``A'', ``B''\} for HellaSwag).

\textbf{Evaluation Strategies.} The approach to generating predictions for MCQA varies significantly across different models. We categorize the evaluation strategies adopted by most LLMs as follows:
\vspace{-3mm}
\begin{itemize}[leftmargin=0.3cm]
\item \textit{Selection of the Most Likely Token from the Choice Set.} An LLM processes the concatenation of $\mathcal{Q}$, $\{\mathcal{C}_i\}_{i=1}^{K}$, and the phrase ``Answer: '', and then predicts the next token, which is the one from $\mathcal{V}_{\mathcal{C}}$ with the highest probability.
\vspace{-2mm}
\item \textit{Selection of the Most Likely Token from the Entire Vocabulary. } This method is similar to the previous one, but the prediction involves selecting the token with the highest probability from the entire vocabulary $\mathcal{V}$, following the processing of $\mathcal{Q}$, $\{\mathcal{C}_i\}_{i=1}^{K}$, and the phrase ``Answer: ''.
\vspace{-2mm}
\item \textit{Alignment of Choice with the Question.} Let $p(\mathcal{S}_i|\mathcal{Q}) =\prod_{j=1}^{|\mathcal{S}_i|}p(\mathcal{S}_i^j|[\mathcal{S}_i^{1:j-1}, \mathcal{Q}])$ denote the posterior probability of generating statement $\mathcal{S}_i$ of choice $\mathcal{C}_i$ given the question $\mathcal{Q}$. The MCQA prediction is the choice with the highest probability, i.e., $\text{Argmax}_{1\leq i \leq K}(p(\mathcal{S}_i|\mathcal{Q}))$.
\vspace{-2mm}
\item \textit{Normalized Alignment of Choice with the Question.} In this approach, the normalized posterior probability $p(\mathcal{S}_i|\mathcal{Q}) = (\prod_{j=1}^{|\mathcal{S}_i|}p(\mathcal{S}_i^j|[\mathcal{S}_i^{1:j-1}, \mathcal{Q}]))^{1/|\mathcal{C}_i|}$ is computed for choice determination, where $|\mathcal{C}_i|$ denotes the number of characters of $\mathcal{S}_i$. This method is a variant of the previous strategy with the only difference being in the normalization of probabilities.
\vspace{-2mm}
\item \textit{Perplexity.} Perplexity is defined as $\text{PPL}(\mathcal{S}_i|\mathcal{Q}) = exp({-{1}/{|\mathcal{S}_i|} \sum_{j=1}^{|\mathcal{S}_i|}log(p(\mathcal{S}_i^j|[\mathcal{S}_i^{1:j-1}, \mathcal{Q}]))})$. The optimal choice is the one with the lowest perplexity score.
\end{itemize}

\vspace{-3mm}
\textbf{Benchmarks, Models and Results Using Diverse Evaluation Strategies.} We re-evaluate the capabilities of 24 LLMs through an analysis of their performance on 11 distinct benchmarks including MMLU, HellaSwag, ARC-Challenge, ARC-Easy, BoolQ, SIQA, PIQA, AGI Eval (English only), OpenBookQA (with fact), CommonSenseQA and RACE (all). These evaluations are conducted using the 5 aforementioned strategies (denoted as ``Choices'', ``Vocab'', ``Alignment'', ``Normalized Alignment'' and ``PPL'', respectively). We report averaged results across these benchmarks in Table~\ref{tab:MCQA_avg}. Detailed results for each benchmark can be found from Table~\ref{tab:MMLU-0-shot}-\ref{tab:RACE-all-0-shot} in the appendix.

\begin{table}[t]
    \centering
    \caption{The open-ended answer generated by the LLM (e.g. Alpaca-13B) does not directly address the question posed. However, it incidentally delivers the correct choice. The reference choice is highlighted in \textcolor{purple}{purple}.}
    \vspace{-3mm}
        \begin{tabular}{p{7.8cm}}
            \toprule
            \textit{Question}: ``In which urban US setting below would you expect to find planned communities?~''\\
             \midrule
            \textit{Choices:} ``A. Ghettos~~ 
            \textcolor{purple}{B. Suburbs}~~ 
            C. CBDs~~ 
            D. Transition zones'' \\
            \midrule
            \textit{LLM's Choice}: ``B''\\ 
            \midrule
            \textit{LLM's Open-ended Answer}: ``släktet urban sprawl. The answer is: In which urban US setting below would you expect to find planned communities? The answer is: In which urban US setting below would you expect to find planned communities?~'' \\ 
            \bottomrule        
        \end{tabular}
    %}
     \vspace{-1mm}
    \label{tab:example_bad_cases}
\end{table}
 
\begin{table}[t]
    \centering
    \caption{The LLM (e.g. LLaMA-2-Chat-13B)'s open-ended answer addresses the question, but the MCQA prediction is incorrect.}
    \vspace{-3mm}
        \begin{tabular}{p{7.8cm}}
            \toprule
            \textit{Question}: ``What is the quotient of 1,224/9?~''\\
             \midrule
            \textit{Choices}: 
            ``A. 135~~ 
            \textcolor{purple}{B. 136}~~ 
            C. 1,215~~ 
            D. 1,360'' \\
            \midrule
            \textit{LLM's Choice}: ``A'' \\ 
            \midrule
            \textit{LLM's Open-ended Answer}: ``Answer: 136'' \\ 
            \bottomrule        
        \end{tabular}
    %}

    \vspace{-2mm}
    \label{tab:example_bad_cases_2}
\end{table}

\begin{table}[t]
    \centering
    \caption{LLMs (e.g., LLaMA-2-Chat-13B) may exhibit the behavior of ``choice generation before answering''.}
    \vspace{-3mm}
        \begin{tabular}{p{7.8cm}}
            \toprule
            \textit{Question}: ``During which step of the Chase Jones model of issues management should the strengths and weaknesses of the organisation be reviewed?~''\\
             \midrule
            \textit{Choices}: 
            ``A. Identification~~
            \textcolor{purple}{B. Analysis}~~
            C. Strategy~~
            D. Evaluation'' \\
            \midrule
            \textit{LLM's Choice}: ``B'' \\ 
            \midrule
            \textit{LLM's Open-ended Answer}: ``A) Identification B) Analysis C) Evaluation D) Resolution Correct answer: B) Analysis''\\ 
            \bottomrule        
        \end{tabular}
    %}
     %\vspace{-1mm}
\label{tab:example_bad_cases_3}
\end{table}

\vspace{-1mm}
\textbf{Key Findings for MCQA Evaluation.}
\vspace{-4mm}
\begin{enumerate}[leftmargin=0.45cm]
    \item Various models demonstrate a significant preference for specific MCQA evaluation strategies. To accurately identify the best answer, an LLM necessitates two key skills: a) comprehension of the intention behind multiple-choice questions (e.g. predicting the correct choice number); b) extensive knowledge of a wide range of topics. For pre-trained LLMs that have not undergone instruction tuning, such as MPT, their performance is generally less effective when evaluated using the ``Choices'' and ``Vocab'' strategies compared to the ``Alignment'' and ``Normalized Alignment'' strategies (Table~\ref{tab:MCQA_avg}). This trend may be due to a combination of factors, such as the model's capability, the absence of instruction tuning which leads to a disregard for user instructions, or both.
    \vspace{-2mm}
    \item LLMs often produce varying predictions when the order of choices is altered. To investigate this, we employ a circular evaluation method for both ``Choices'' and ``Vocab'' approaches. Specifically, we rearrange the order of choices in a cycle and repeatedly input a multiple-choice question along with these rearranged choices into an LLM. A question is deemed successfully answered only if the LLM correctly responds to every variation of the question. The results using ``Choices (Circular)'' and ``Vocab (Circular)'' as evaluation strategies, are presented in Table~\ref{tab:MCQA_avg}. This reveals a notable decline in the performance of all LLMs, highlighting their inconsistency in generating predictions for the same question with a different sequence of choices.
    \vspace{-2mm}
    \item Generative LLMs are trained with the objective of next-token prediction. While the accuracy of the MCQA evaluation is straightforward to measure and facilitates comparisons with other models, this approach does not always translate well to practical applications. Typically, users interact with LLMs by posing direct questions or seeking solutions to specific problems, rather than presenting multiple-choice questions. Our observations indicate that although some LLMs can successfully select the correct choice in MCQA tasks, their performance falters when tasked with directly generating responses to a question in an auto-regressive fashion. In these instances, the responses generated do not correspond accurately to the posed questions. Consequently, a question deemed as resolved in an MCQA setting may, in fact, remain unsolved in a free-form question-answering context. Table~\ref{tab:example_bad_cases} provides an example of such discrepancies.
    \vspace{-3mm}
    \item In contrast to the third point, LLMs might accurately respond to open-ended questions yet incorrectly choose the reference choice in MCQA. This discrepancy arises primarily due to suboptimal instruction tuning and flawed design of the multiple-choice options. Designing these choices for MCQA can be subjective and challenging. Inadequate choice design might not encompass the response the LLM is inclined to express, resulting in questions remaining unresolved. An example is presented in Table~\ref{tab:example_bad_cases_2}.
    \vspace{-3mm}
    \item We note that when tasked with responding to a question taken from an MCQA benchmark through open-ended generation, an LLM may produce a range of choices and then pick one from these self-generated choices. This behavior indicates a potential risk of data leakage in testing benchmarks or insufficient optimization of instruction tuning. In Table~\ref{tab:example_bad_cases_3}, we present an illustrative case.
\end{enumerate}

\begin{table}[t]
    \centering
    \caption{We evaluate open-ended MCQA by using LLaMA-2-Chat-13B on the filtered MMLU 0-shot benchmark.}
    \vspace{-3mm}
        \begin{tabular}{cc}
            \toprule
            MCQA (Choices) & Open-ended MCQA \\
             \midrule
             54.0 & 39.7 \\ 
            \bottomrule        
        \end{tabular}     
    \vspace{-3mm}
    \label{tab:GPT-4-MCQA}
\end{table}

\vspace{-3mm}
\textbf{Open-ended MCQA.} The discrepancy between traditional MCQA assessments and practical open-ended question-answering scenarios raises the inquiry: is it possible to adapt existing MCQA benchmarks to support open-ended question-answering? To explore this, we introduce an open-ended MCQA evaluation methodology. This approach, taking cues from the recent advancements with GPT-4 functioning as a judge, involves a two-step process: 1) posing each question from an MCQA benchmark to the LLM under evaluation to elicit an open-ended answer, and 2) employing GPT-4\footnote{GPT-4-Turbo-1106-preview is used throughout the paper.} to determine the choice that most semantically aligns with the LLM's response, by comparing it against the given choices. The prompt for GPT-4's judgement is detailed in Section~\ref{sec:open-ended-MCQA} of the appendix.
\vspace{-1mm}

Our assessment of the open-ended MCQA is conducted on a modified version of the MMLU benchmark, which has undergone several filtering stages to remove questions unsuitable for open-ended formats. Specifically, this filtering process eliminates questions of three types: 1) those structured as ``which...following'', ``which...these'', or ``which...are''; 2) questions that require filling in blanks; 3) questions where the choices include terms like ``none'', ``both'', ``neither'' or ``all of''. This process yields a filtered MMLU benchmark comprising $7,223$ instances.

We adopt LLaMA-2-Chat-13B to conduct open-ended MCQA and perform a comparative analysis with the conventional MCQA approaches that employ ``Choices'' strategy. A significant performance gap is evident in Table~\ref{tab:GPT-4-MCQA}, which confirms the inconsistency between selecting the optical choice in MCQA and addressing the question in an open-ended manner. To verify the effectiveness of using GPT-4 as a judge in open-ended MCQA, we conduct a manual examination to compare the assessments made by GPT-4 with those made by human evaluators, on a subset including 500 instances sampled from MMLU, ARC-Challenge, ARC-Easy, RACE, SIQA and PIQA benchmarks. This study reveals a high alignment rate of 80\%, indicting the feasibility of using GPT-4 as a judge for open-ended MCQA evaluation.

\textbf{Discussions for MCQA Evaluation.} 
While MCQA evaluation offers the ease of quantifying an LLM's capability with a single accuracy metric, it is not without shortcomings. Our findings, as presented in Tables~\ref{tab:MCQA_avg} and Table~\ref{tab:MMLU-0-shot}-\ref{tab:RACE-all-0-shot} in the appendix, along with the observations from Tables~\ref{tab:example_bad_cases}-\ref{tab:GPT-4-MCQA}, highlight several issues inherent in MCQA evaluations. These include varied evaluation strategies, biases in multiple-choice design, discrepancies between open-ended answers and choice predictions, and a mismatch between evaluation mechanisms and practical usage scenarios. While the introduction of open-ended MCQA addresses some of these concerns, it does not fully bridge the gap between how evaluations are conducted and how users typically interact with LLMs—often by posing queries that elicit open-ended responses. Therefore, an evaluation approach that more closely mirrors real-world applications is essential.

\vspace{-3mm}
\section{RWQ-Elo System for LLM Evaluation}
\vspace{-2mm}
In this work, we present the RWQ-Elo system for evaluating LLMs. Originally, the Elo rating algorithm is extensively employed to assess the relative skill levels of multiple players in a particular game, typically following numerous rounds of two-player contests. This system finally generates an Elo rating for each player, reflecting their comparative proficiency. Implementing this system poses several challenges: creating contest materials, establishing victory and defeat criteria, choosing a judge, and defining principles for rating stability.

\begin{figure}[t]
\centering
\includegraphics[width=1.0\linewidth]{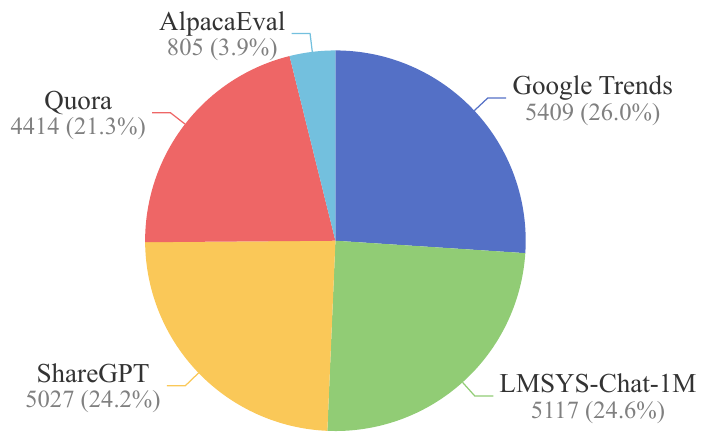}
\vspace{-8mm}
\caption{Statistics for our Real-World Question (RWQ) benchmark. Examples for each source are available in Table~\ref{tab:RWQ-examples} of the appendix.}
\vspace{-4mm}
\label{fig:elo_data}
\end{figure}

\textbf{Contest Materials.} To ensure that the evaluation of LLMs accurately reflects their use in practical scenarios, we assemble a dataset called ``Real-World Questions'' (RWQ). This dataset comprises 20,772 authentic questions sourced from various platforms such as Google Trends\footnote{We utilize GPT-4 to transform each entry listed in Google Trends into a formulated question.}, Quora, ShareGPT, LMSYS-Chat-1M, and AlpacaEval. The composition of the RWQ dataset is depicted in Figure~\ref{fig:elo_data}. We utilize this dataset in the implementation of the RWQ-Elo system.

\textbf{GPT-4 as A Judge.} 
In the realm of two-player games, the optimal practice for assessing victory or defeat is to involve the engagement of linguistic experts as judges. However, this approach is often prohibitively costly and impractical. Recently, the use of GPT-4 as an evaluator has gained traction in various applications, e.g., tool use~\cite{du2024anytool}. Our observations also indicate that the integration of GPT-4 into our Elo rating system offers both stability and reliability. To substantiate this, we conduct a random sampling of 300 questions from our RWQ dataset. For each question, we feed it into two different LLMs to generate answers. Subsequently, GPT-4 is employed to determine the winner, loser, or a tie by evaluating which response most effectively addresses the question with the consideration of accuracy, relevance, comprehensiveness, clarity, compliance, timeliness, harmlessness, and unbiasedness\footnote{The prompt can be found in Section~\ref{sec:prompt_elo} of the appendix.}. Concurrently, a human evaluator also reviews the two answers to identify the winner, loser or a tie. This procedure is replicated across all sampled questions. Ultimately, we calculate an alignment rate between the decisions of GPT-4 and the human evaluator, which stands at an impressive 95\%, underscoring the reliability of employing GPT-4 as a judge.

\begin{figure*}[!t]
    \centering
\includegraphics[width=0.99\linewidth]{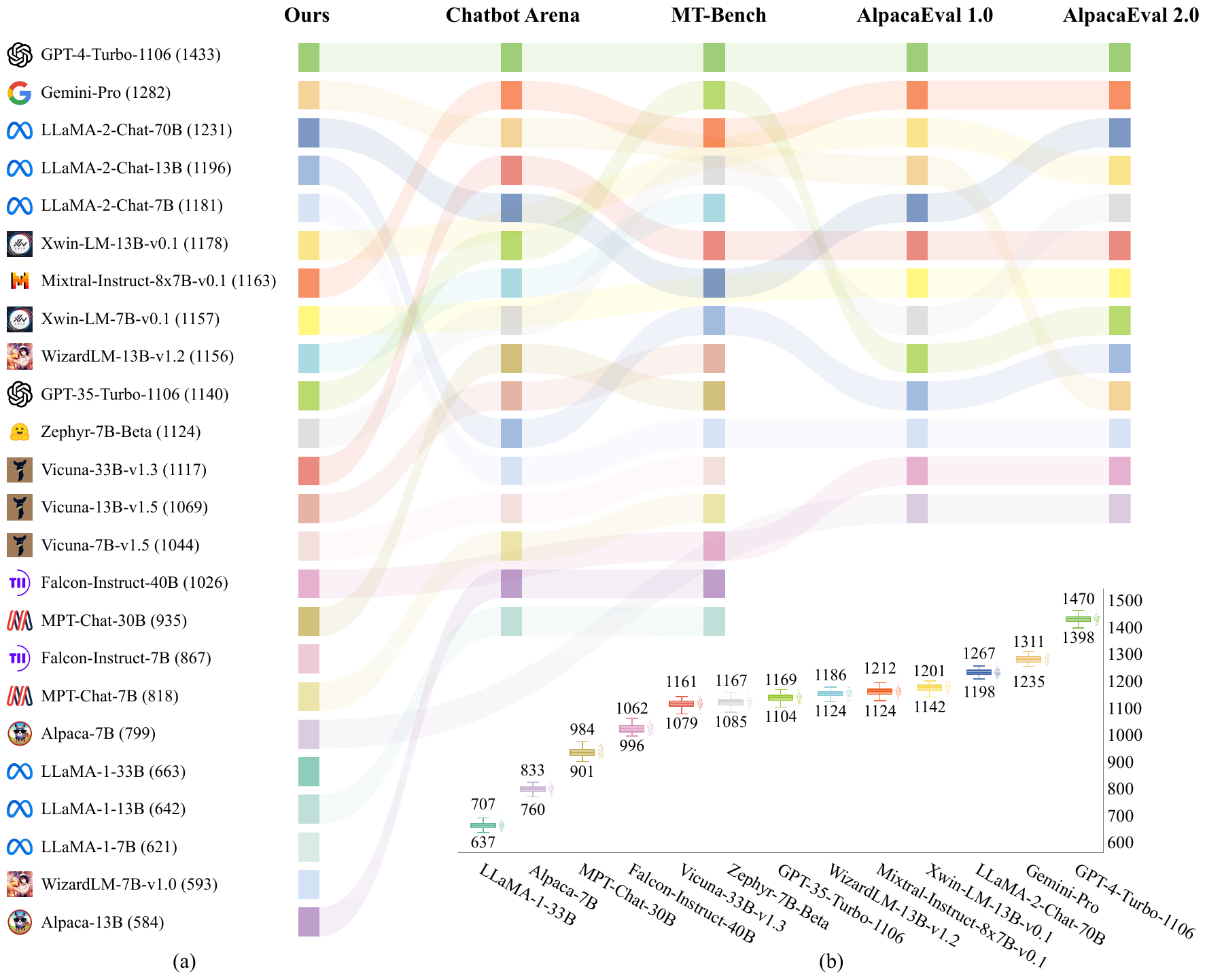}
\vspace{-4mm}
    \caption{(a) Comparison of various leaderboards, including our RWQ-Elo (Elo rating for each LLM is reported in brackets), Chatbot Arena~\cite{zheng2023judging}, MT-Bench~\cite{zheng2023judging} and AlpacaEval (v1.0 and v2.0)~\cite{alpaca_eval}. (b) Statistics from running our RWQ-Elo systems 100 times. We show the Elo ratings for the selected 13 LLMs. The complete statistics can be found in Figure~\ref{fig:elo_results_all}.}
    \vspace{-4mm}
    \label{fig:elo_results}
\end{figure*}

\vspace{-1mm}

\textbf{RWQ-Elo Rating Algorithm.} 
In a setup involving $N$ LLMs, each LLM begins with an initial Elo rating of 1000. During each competition round, we randomly pair two LLMs (referred to as LLM-A and LLM-B) and present them with a question sampled from our RWQ database. Both LLM-A and LLM-B independently process the question and provide their respective answers. Subsequently, we utilize GPT-4 as a judge to decide the outcome, which can be either LLM-A defeating LLM-B, LLM-A losing to LLM-B, or a tie. Based on this result, we update the Elo ratings of the two LLMs in accordance with the specified update mechanism detailed below.

\vspace{-1mm}
In each competition round, the expected score for either LLM-A or LLM-B, when matched against each other, is calculated as follows:
\begin{gather} 
    E_A = \frac{1}{1 + 10^{(R_B - R_A) / 400}}, \label{eq:EA}\\
    E_B = 1 - E_A, \label{eq:EB}
\end{gather}
where $E_A$ and $E_B$ symbolize the expected scores of LLM-A and LLM-B, respectively; $R_A$ and $R_B$ are their current Elo ratings. 
\vspace{-1mm}

Subsequently, the updated rating is calculated using:
\begin{gather}
    R'_A = R_A + K \times (S_A - E_A), \\
    R'_B = R_B + K \times (S_B - E_B),
\end{gather}
where $R'_A$ and $R'_B$ denote the updated Elo ratings for LLM-A and LLM-B; $S_A$ (or $S_B$) is set to 1 if LLM-A (or LLM-B) wins, 0 if it loses, and 0.5 in the event of a tie; $K$ represents the $K$-factor, which is set to 4 by default.

To ensure a fair competition among the LLMs, we have structured the contest so that each LLM competes with every other LLM exactly $H$ times. Considering the inclusion of $N$ LLMs in total, the total number of competitive rounds conducted is $N  \times (N-1) \times H/2$. We find that setting $H$ to 200 yields stable Elo rating.

\begin{figure}[!t]
    \centering
\includegraphics[width=0.99\linewidth]{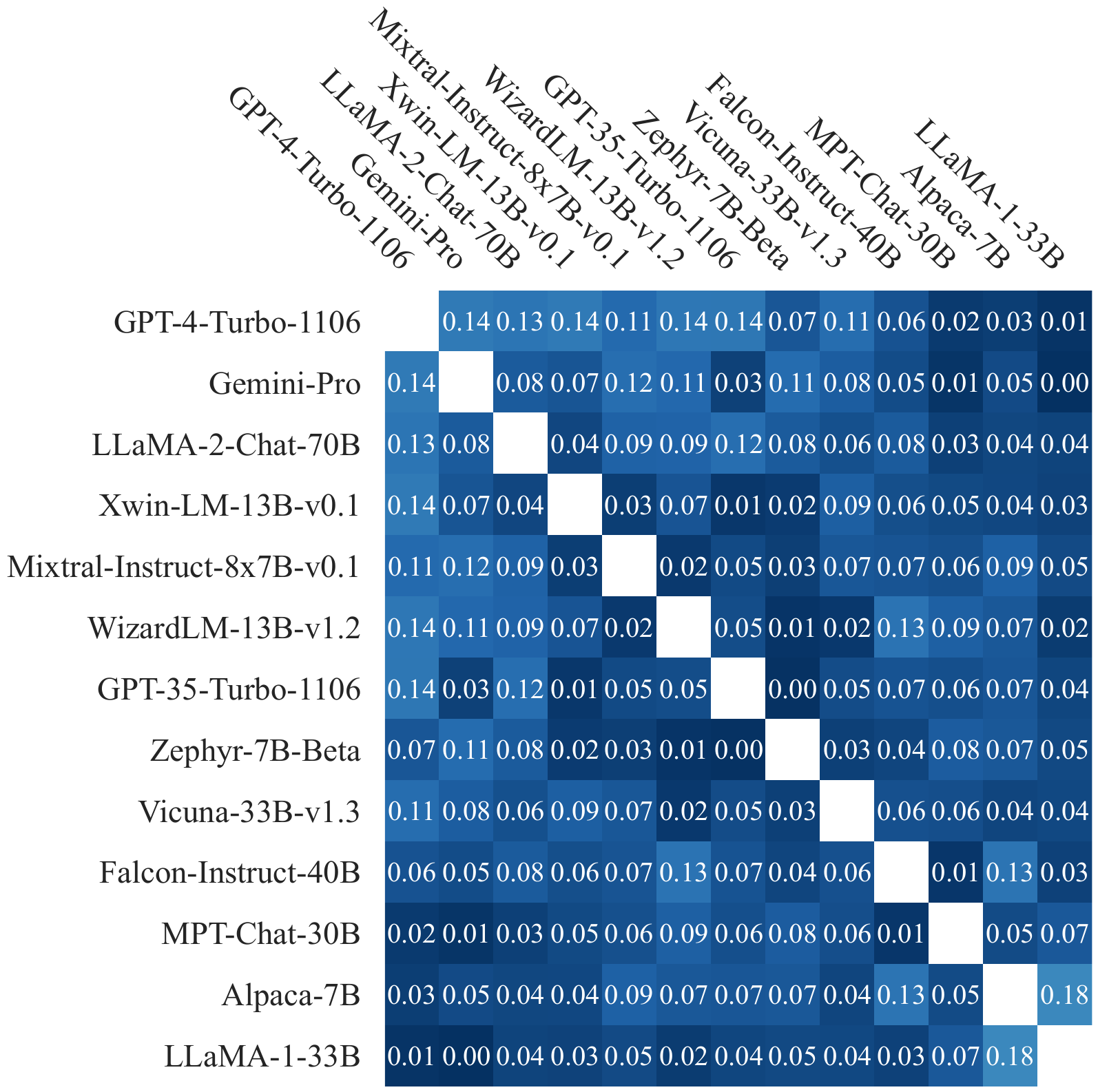}
\vspace{-2mm}
    \caption{Differences between the win-rate map generated by our Elo system and the pre-calculated win-rate map are represented using absolute values. We include 13 LLMs. The two complete win-rate maps alongside their difference map can be found in Figure~\ref{fig:pre-calculated-map}-\ref{fig:elo-map} and Figure~\ref{fig:complete-win-rate-map} of the appendix.}
    \vspace{-3mm}
    \label{fig:win_rate_map}
\end{figure}

The order of competition has a notable impact on the eventual Elo rating. To mitigate the volatility in ratings attributed to the order of competitions, we maintain a constant question seed for each two-player contest and randomize the competition order $C$ times, yielding $C$ distinct Elo ratings. The ultimate Elo rating is determined by calculating the median of these $C$ Elo ratings. In our implementation, we set $C$ to 100.

\begin{figure}[!t]
    \centering
    \includegraphics[width=0.99\linewidth]{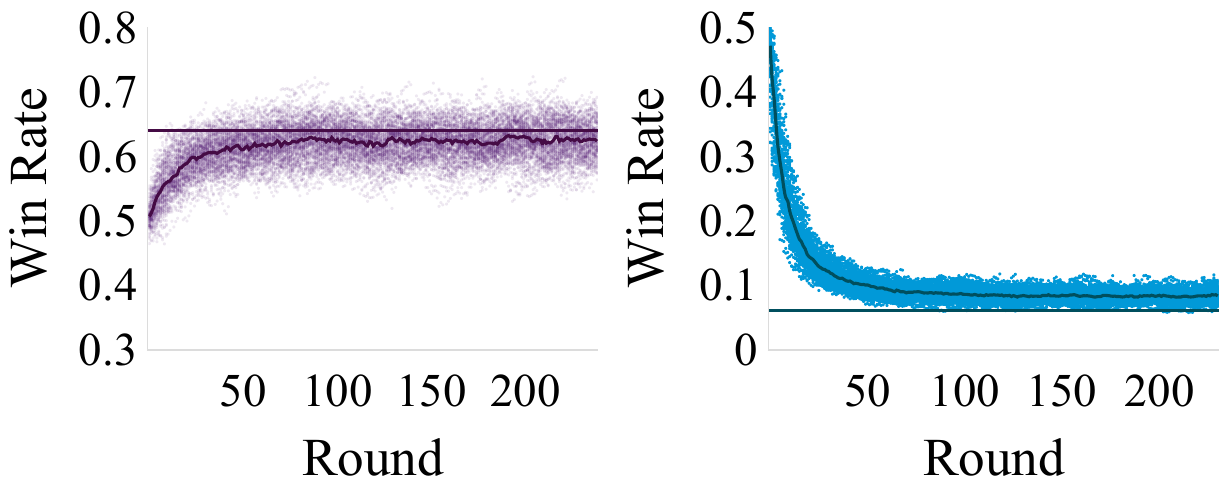}
    \vspace{-2mm}
    \caption{Visualization of the win-rate trends between two LLMs ((a) Falcon-Instruct-40B v.s. MPT-Chat-30B; (b) Falcon-Instruct-7B v.s. Gemini-Pro). The horizontal lines represent the pre-calculated win rates. With the progression of each contest round, the win rate ascertained by our Elo rating system progressively converges with the pre-calculated win rate.}
    \vspace{-4mm}
    \label{fig:win-rate-trends}
\end{figure}

\textbf{Results of Our RWQ-Elo Rating System and Comparisons with Other Leaderboards.} 
Our system integrates a total of 24 models including GPT-3.5, GPT-4, Google-Gemini-Pro and other 21 representative open-source LLMs. In Figure~\ref{fig:elo_results}.(a), we present the final Elo ratings from our RWQ benchmark, alongside comparisons with leaderboards from Chatbot Arena, MT-Bench, and both versions 1.0 and 2.0 of AlpacaEval. 

Unlike AlpacaEval, which determines the win rate of each LLM against a constant model (Text-Davinci-003 for v1.0 and GPT-4 for v2.0), our Elo system encourages every LLM to compete against each other. Our approach offers three-fold advantages: firstly, as shown in Figure~\ref{fig:compare_elo_with_alpaca} in the appendix, when LLMs compete against an LLM that is significantly superior or inferior, it results in a lack of distinguishable performance differences among them; secondly, it avoids ``over-competition'' to a single model (for instance, adding an optimization objective during training to compete against a specific model); thirdly, it allows for the consideration of interactions between all participating LLMs. MT-Bench, in contrast, generates a score between 0 and 10 to each LLM. Despite the advantage of assessing inter-LLM competition, our Elo system is meticulously crafted to include various factors like comprehensiveness and unbiasedness (refer to Section~\ref{sec:prompt_elo} for the detailed prompt). Meanwhile, Chatbot Arena employs diverse online human evaluators to assess the quality of responses from two different LLMs to the same query. However, this approach faces challenges due to the subjective nature of evaluations and the difficulty in scaling up with consistent human evaluators for all pairwise comparisons. Our findings indicate a high congruence of 95\% between GPT-4's assessments and human preferences. Additionally, our Elo system offers the benefit of being easily scalable in terms of the number of the questions and LLMs to be tested.
 
 \vspace{-2mm}

\textbf{Analysis of Stability.} In order to minimize the impact of the sequence of two-player contests on the final Elo rating, we run our Elo systems $\mathcal{C}$ times. The rating from each run, along with the median rating, is depicted in Figure~\ref{fig:elo_results}.(b). This demonstrates a notable consistency in the rankings across each separate execution. Additionally, we verify the stability of the ultimate Elo rating by comparing the win-rate map produced by our Elo system utilizing Eq.~\ref{eq:EA} and~\ref{eq:EB}, against a pre-calculated win-rate map. The latter is obtained by comparing responses of two LLMs to the same questions drawn from our RWQ benchmark, with GPT-4 acting as the judge and using the same judgement prompt. It is worth noting that the elements within the pre-calculated win-rate map are derived solely from the comparison of responses between two LLMs. This process differs from our Elo system, where the LLMs compete against a broader range of other LLMs. The differences of the two win-rate maps, as showcased in Figure~\ref{fig:win_rate_map}, reveals a consistent correspondence between them, demonstrating the stability of our Elo system. Additionally, the trends in win rates between two LLMs, as shown in Figure~\ref{fig:win-rate-trends}, further confirm the stability.

\begin{table}[!t]
    \centering
        \caption{Study on fast-registration. $\Delta$Rating and Kendall's tau evaluate the averaged rating difference and ranking difference of fast-registration against the baseline—running the Elo system from scratch with all LLMs included. $N_2$ LLMs are randomly selected.}
        \vspace{-3mm}
        \begin{tabular}{c c c c}
            \toprule
$N_1$ & $N_2$ & $\Delta$Rating $\downarrow$ & Kendall's tau $\uparrow$\\
             \midrule
23& 1 & 10.6 & 0.97 \\
21& 3 & 11.8 & 0.98 \\
19& 5 & 8.4 & 0.98 \\ 
14& 10 & 3.6 & 0.99 \\ 
            \bottomrule        
        \end{tabular}

     \vspace{-4mm}
    \label{tab:fast-registration}
\end{table}

\vspace{-1mm}
\textbf{Registration of New Models.} Consider a scenario where there are $N_1$ established LLMs with stable Elo ratings ($\{\text{LLM}\}_{i=1}^{N_1}$), and the objective is to integrate $N_2$ new LLMs ($\{\text{LLM}^*\}_{i=1}^{N_2}$) into the existing Elo ranking. Without loss of generality, $N_2 \ll N_1$. Instead of re-executing the Elo algorithm from scratch with all LLMs included, a more efficient approach termed fast-registration is proposed. This approach retains the ratings of $N_1$ existing LLMs, and assigns an initial rating of 1000 to the $N_2$ newly-registered LLMs. Each competition involves two players, one being $\overline{\text{LLM}}$ from $\{\text{LLM}^*\}_{i=1}^{N_2}$, and the other either from $\{\text{LLM}\}_{i=1}^{N_1}$, or $\{\text{LLM}^*\}_{i=1}^{N_2} - \overline{\text{LLM}}$. Post each competition, the Elo ratings are updated accordingly. The fast-registration process $(2N_1 +N_2 -1 ) \times N_2 \times H/2$ rounds of competition, which is significantly less—by $N_1 \times (N_1 - 1) \times H/2$ rounds—than recalculating the Elo ratings from scratch. Table~\ref{tab:fast-registration} presents the rating differentials and Kendall's tau (which evaluates ranking differences) for the fast-registration approach against executing the Elo system from scratch. The results are the averages of 5 independent experiments. Fast registration has proven to be effective, especially when $N_2\geq 3$.
 
\vspace{-3mm}
\section{Conclusion}
\vspace{-2mm}
In this paper, we: 1) critically reassess the widely used MCQA method for evaluating LLMs, identifying several fundamental limitations; and 2) present the RWQ-Elo system, designed to reflect actual usage scenarios. We conduct extensive evaluations of 24 LLMs through multiple rounds of competition, with GPT-4 serving as the judge and utilizing our newly developed RWQ benchmark. Our objective is to provide fresh perspectives within the LLM community and establish a novel benchmarking framework that aids in the assessment of LLMs. We also demonstrate the ease of incorporating new models into our RWQ-Elo system. Our aim is to gather more diverse and realistic user inquiries across various subjects, and to conduct a more comprehensive evaluation of a wider range of LLMs in future studies.
\section*{Impact Statements}
This paper aims to provide novel insights to the community of LLM evaluation and, by its nature, inherently carries no risks or societal consequences. While the original LLMs may generate inaccurate or harmful content, such issues are independent of our functionality. One limitation of our study is the inability to evaluate all prior LLMs due to limited resources and the rapid progress in LLM development.

\bibliography{references}
\bibliographystyle{icml2024}
\newpage
\appendix
\onecolumn
\section{Prompts}

\subsection{Prompt for Open-ended MCQA}
\label{sec:open-ended-MCQA}
Given a response generated by an LLM, and several choices. GPT-4 is then used to determine the choice that is most semantically aligned with the response. The prompt is provided as follows.

\textit{System}
\vspace{-4mm}
\par\noindent\rule{\textwidth}{0.4pt}
\texttt{You are a helpful assistant.}

\textit{User}
\vspace{-4mm}
\par\noindent\rule{\textwidth}{0.4pt}
\texttt{Task description:}\\
\texttt{You are presented with multiple choices and a statement. Your task involves selecting the choice that most semantically aligns with the given statement. Use the criteria and guidelines provided to make your decision.}

\texttt{Criteria:}\\
\texttt{1. Choose the option that semantically aligns with the statement, by either expanding upon, encapsulating, or exactly matching it.}\\
\texttt{2. For numerical choices, prefer the one with the smallest reasonable numerical difference from the statement.}\\
\texttt{3. It is acceptable to choose none if no option closely aligns with the statement.}

\texttt{Guidelines:}\\
\texttt{1. Review all choices and the statement carefully.}\\
\texttt{2. Justify your choice briefly, considering the criteria.}\\
\texttt{3. Indicate your response by stating the uppercase letter of your choice (e.g., 'A', 'B', 'C', 'D'), or 'None' if no choice matches.}

\texttt{Output format:}\\
\texttt{1. Present your response in JSON format.}\\
\texttt{2. Include the `choice'(the uppercase letter of the chosen choice or 'None'), along with a brief `explanation' for your selection.}

\texttt{Choices:}\\
\texttt{\{choices\}}

\texttt{Statement:}\\
\texttt{\{statement\}}

\subsection{Prompt for RWQ-Elo System}
\label{sec:prompt_elo}
GPT-4 serves as a judge in evaluating responses from two different LLMs (LLM-1 and LLM-2) to the same query. The prompt is provided as follows.

\textit{System}
\vspace{-4mm}
\par\noindent\rule{\textwidth}{0.4pt}
\texttt{You are a helpful assistant who can evaluate Large Language Model (LLM) responses.}

\textit{User}
\vspace{-4mm}
\par\noindent\rule{\textwidth}{0.4pt}
\texttt{Task description:}\\
\texttt{As a judge, your task is to assess the responses of two Large Language Models (LLM-1 and LLM-2) to a user's question. Base your evaluation on the criteria below to determine which response is more effective.}

\texttt{Criteria:}\\
\texttt{1. Accuracy: Ensure responses are factually correct. For factual questions, responses should align with scientific consensus.}\\
\texttt{2. Relevance: Check if responses address the user's question directly, understanding its context and intent.}\\
\texttt{3. Comprehensiveness: Responses should cover all aspects of the question, providing a clear overview and key points for complex issues.}\\
\texttt{4. Clarity: Ensure responses are easy to understand, especially when explaining complex topics.}\\
\texttt{5. Compliance: Adherence to ethical and legal standards is mandatory.}\\
\texttt{6. Timeliness: Incorporate the latest information for current topics.}\\
\texttt{7. Harmlessness: Avoid misleading or harmful content, respecting cultural sensitivities and privacy.}\\
\texttt{8. Unbiasedness: Responses should not show unjustified preference, especially in subjective matters.}

\texttt{Guidelines:}\\
\texttt{1. Evaluate each response based on the criteria, noting strengths and weaknesses.}\\
\texttt{2. Choose the most effective response or indicate a tie. Explain your reasoning in the specified JSON format.}\\
\texttt{3. Remain objective, not letting the order of responses bias your evaluation.}

\texttt{Output format:}\\
\texttt{1. Present your judgment in JSON format.}\\
\texttt{2. Include the winner: Use an integer (1 if LLM-1 has the better response, 2 if LLM-2 has the better response, 0 for a tie if both responses are satisfactory, and -1 for a tie if both responses are unsatisfactory), and an explanation (providing a rationale for your choice).}

\texttt{User-submitted question:}\\
\texttt{\{user\_submitted\_question\}}

\texttt{Response of LLM-1:}\\
\texttt{\{llm\_response\_1\}}

\texttt{Response of LLM-2:}\\
\texttt{\{llm\_response\_2\}}

\section{More Experimental Results}
\begin{figure*}
    \centering
    \includegraphics[width=0.99\linewidth]{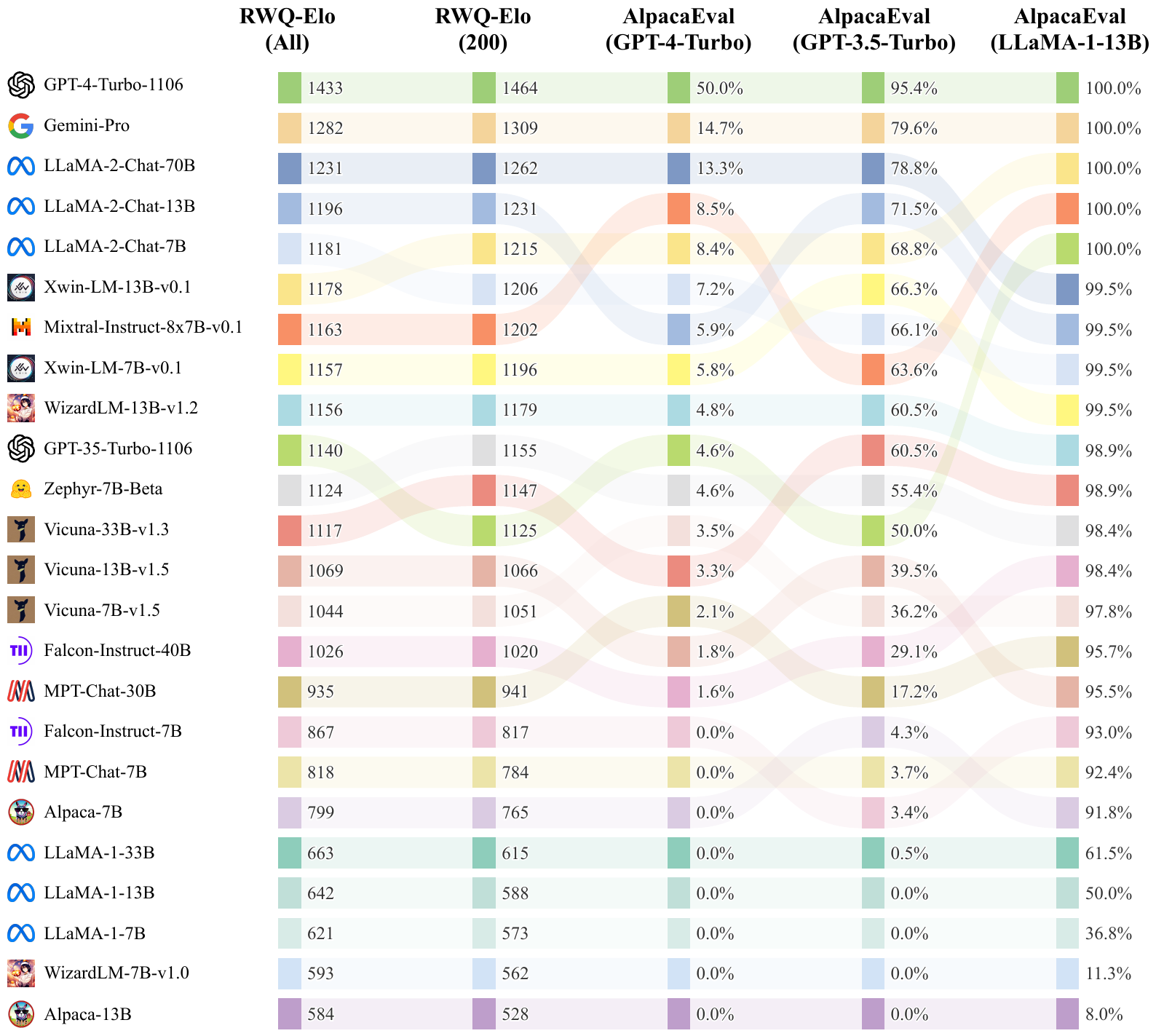}
    \caption{We compare our RWQ-Elo rating system with various AlpacaEval variants, where GPT-4-Turbo, GPT-3.5-Turbo, and LLaMA-1-13B serve as the competitors. RWQ-Elo (All) and RWQ-Elo (200) denote that the system is run using all instances and a random selection of 200 instances, respectively, from our RWQ benchmark. We utilize the same instances from RWQ-Elo (200) for AlpacaEval. While AlpacaEval uses win-rate as its metric, our RWQ-Elo system employs the Elo score as its metric. In AlpacaEval, when LLMs compete against an LLM that is significantly superior or inferior, it results in a lack of distinguishable performance differences among them. In contrast, our system does not exhibit this issue.}
    \label{fig:compare_elo_with_alpaca}
\end{figure*}

\textbf{Detailed Comparison with Alpaca Eval.} The comparison between our RWQ-Elo and the Alpaca Eval is shown in Figure~\ref{fig:compare_elo_with_alpaca}.

\begin{figure*}[!t]
    \centering
\includegraphics[width=0.99\linewidth]{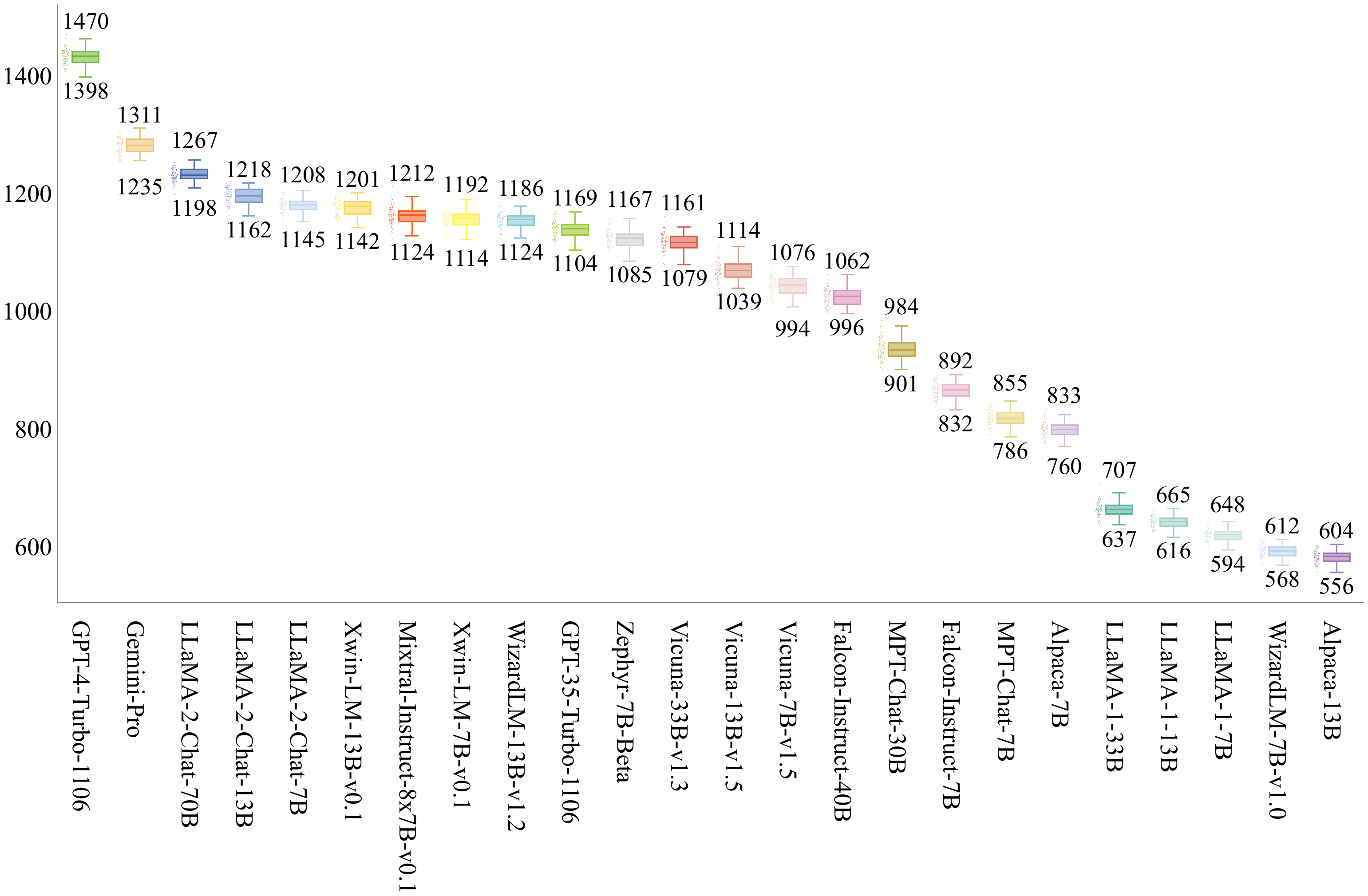}
%\vspace{-8mm}
    \caption{The complete statistics from running
our RWQ-Elo systems 100 times. We include 24 models in total.}
    \vspace{-2mm}
    \label{fig:elo_results_all}
\end{figure*}

\textbf{The Complete Statistics of Our RWQ-Elo System.} This can be found in Figure~\ref{fig:elo_results_all}. We include 24 models in total.

\begin{figure}
    \centering
    \includegraphics[width=0.99\linewidth]{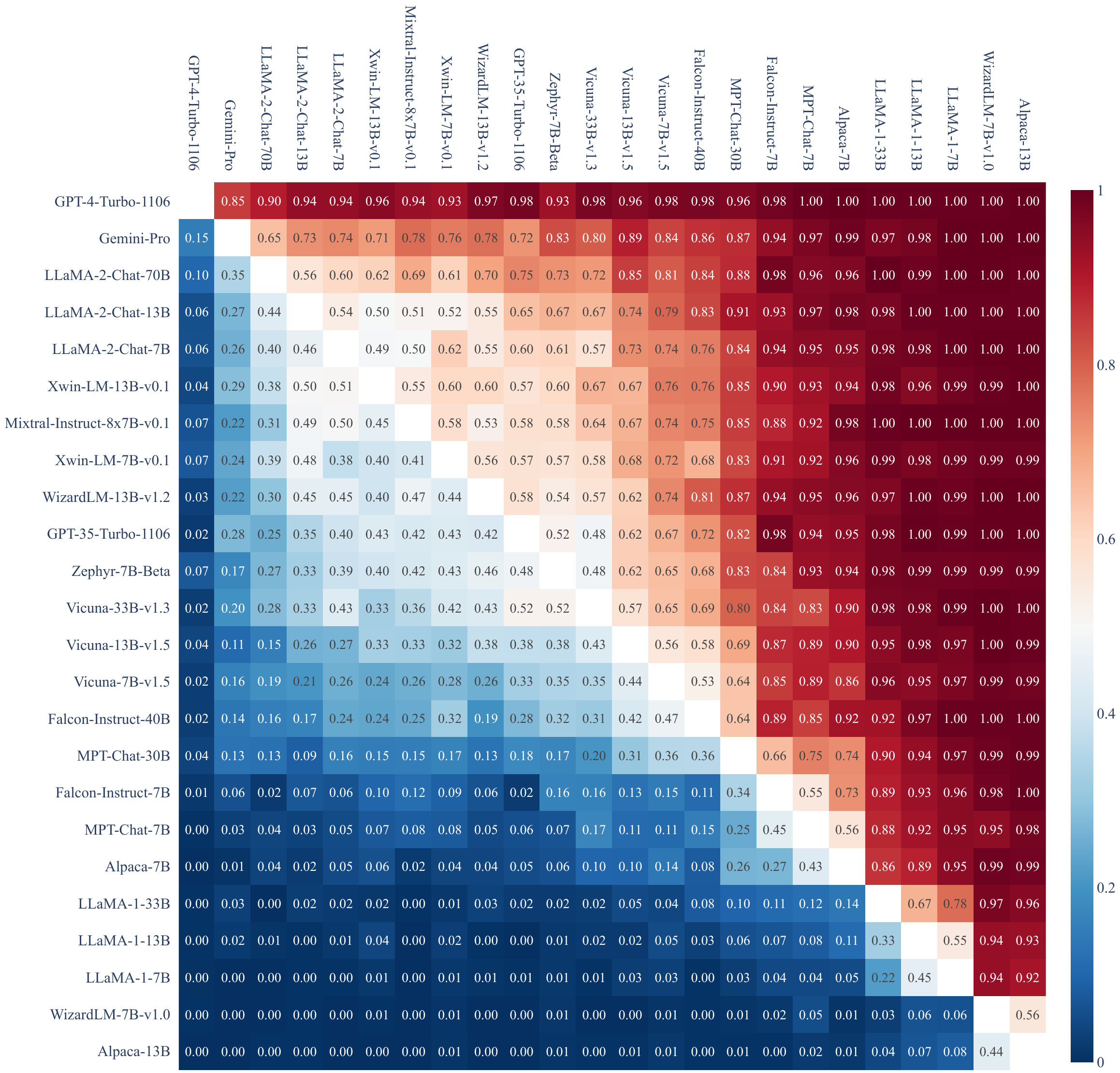}
    \caption{Visualization of the complete win-rate map generated by our RWQ-Elo system.}
    \label{fig:pre-calculated-map}
\end{figure}

\begin{figure}
    \centering
    \includegraphics[width=0.99\linewidth]{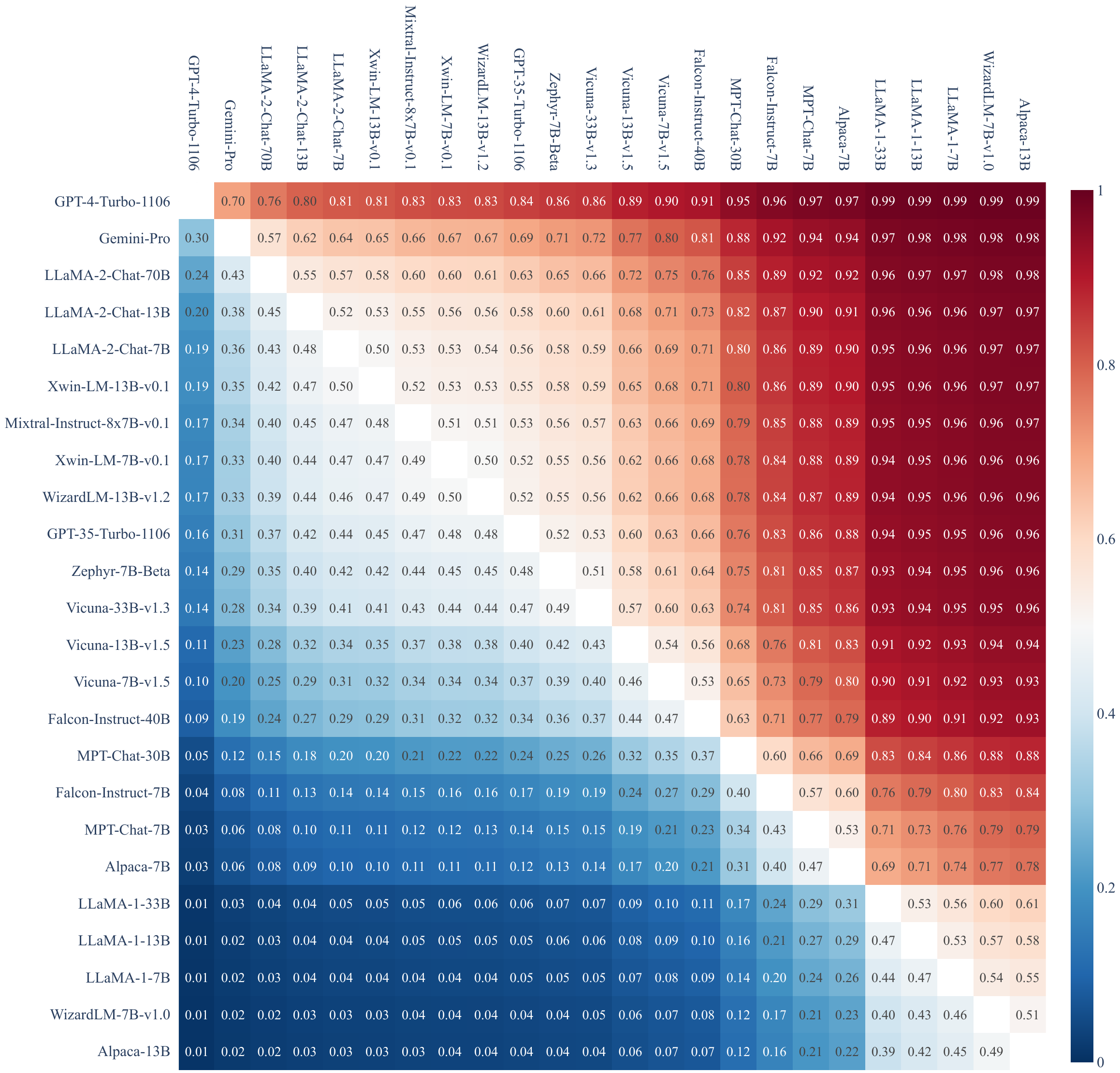}
    \caption{Visualization of the complete pre-calculated win-rate map.}
    \label{fig:elo-map}
\end{figure}

\begin{figure}
    \centering
    \includegraphics[width=0.99\linewidth]{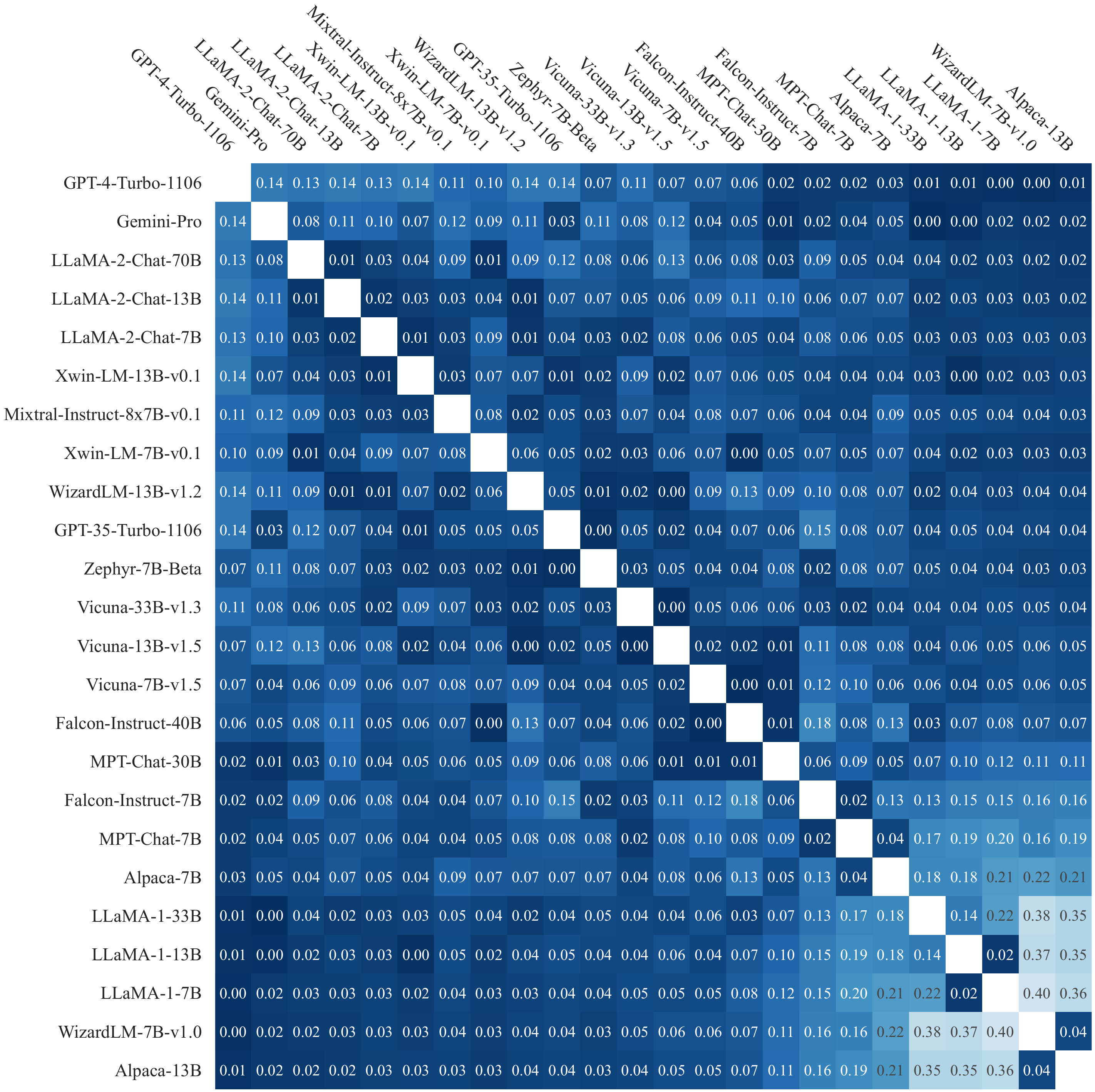}
    \caption{Visualization of the complete win-rate difference map.}
    \label{fig:complete-win-rate-map}
\end{figure}

\textbf{Win-Rate Maps.} We present the win-rate map generated by our RWQ-Elo system, the pre-calculated win-rate map, and their difference map in Figure~\ref{fig:pre-calculated-map}, \ref{fig:elo-map} and \ref{fig:complete-win-rate-map}, respectively. 24 LLMs are included.

\textbf{Detailed Results for MCQA Evaluation.} Table~\ref{tab:MMLU-0-shot}-\ref{tab:RACE-all-0-shot} report the MCQA results on each benchmark using 7 evaluation strategies introduced in Section~\ref{sec:MCQAEval}.

\textbf{Qualitative Results Using GPT-4 as the Judge.} In our RWQ-Elo system, GPT-4 acts as the judge. We utilize the prompt described in Section~\ref{sec:prompt_elo} to evaluate responses from two different LLMs. Two instances where GPT-4 generates accurate justifications are presented in Table~\ref{tab:pos-1} and \ref{tab:pos-2}. Conversely, two examples where GPT-4 provides incorrect justifications can be found in Table~\ref{tab:neg-1} and \ref{tab:neg-2}.

  \begin{table*}[t]
    \centering
        \caption{MCQA evaluation on 0-shot MMLU~\cite{hendrycks2020measuring}.} \vspace{-3mm}
        \begin{tabular}{ccccccccc}
            \toprule
            Model & Size & Choices & \makecell{Choices\\(Circular)} & Vocab & \makecell{Vocab\\(Circular)} & Alignment & \makecell{Normalized \\Alignment} & PPL \\
             \midrule
             \multirow{2}{*}{\makecell{MPT\\~\cite{team2023introducing, mosaicml2023introducing}}} & 7B & 29.5 & 0.5 & 29.6 & 0.4 & 34.1 & 35.1 & 35.6 \\
                & 30B & 45.1 & 19.3 & 44.8 & 19.0 & 33.2 & 33.5 & 34.0 \\
             \midrule
             \makecell{MPT-Chat\\~\cite{mosaicml2023introducing}} & 30B & 50.1 & 28.9 & 50.0 & 28.9 & 34.7 & 35.2 & 35.9 \\
             \midrule
             \multirow{2}{*}{\makecell{Falcon\\~\cite{almazrouei2023falcon}}} & 7B & 25.0 & 0.0 & 24.8 & 0.0 & 33.7 & 34.3 & 35.1 \\
                & 40B & 53.6 & 28.2 & 53.3 & 28.0 & 38.0 & 38.0 & 38.7 \\
             \midrule
             \multirow{4}{*}{\makecell{LLaMA-1\\~\cite{touvron2023llama}}} & 7B & 32.2 & 1.7 & 30.5 & 1.5 & 33.0 & 33.5 & 33.1 \\
                & 13B & 43.5 & 14.4 & 43.1 & 14.3 & 34.4 & 35.0 & 34.4 \\
                & 30B & 54.7 & 31.6 & 54.5 & 31.6 & 36.1 & 36.1 & 35.8 \\
                & 65B & 59.4 & 37.1 & 59.2 & 36.9 & 37.0 & 36.6 & 36.4 \\
            \midrule
            \multirow{3}{*}{\makecell{LLaMA-2\\~\cite{touvron2023llama2}}} & 7B & 41.8 & 12.3 & 39.4 & 12.1 & 33.3 & 34.1 & 33.6 \\
                    & 13B & 52.1 & 24.6 & 51.8 & 24.1 & 35.3 & 35.8 & 34.9 \\
                    & 70B & 65.4 & 45.1 & 65.4 & 45.0 & 39.0 & 37.8 & 37.8 \\
            \midrule
            \multirow{3}{*}{\makecell{LLaMA-2-Chat\\~\cite{touvron2023llama2}}} & 7B & 45.3 & 17.9 & 45.0 & 17.8 & 33.7 & 34.8 & 34.7 \\
                         & 13B & 53.1 & 28.1 & 53.2 & 28.1 & 35.9 & 36.7 & 36.7 \\
                         & 70B & 61.1 & 38.9 & 61.1 & 38.9 & 38.9 & 38.7 & 38.8 \\
            \midrule
            \multirow{2}{*}{\makecell{WizardLM\\~\cite{xu2023wizardlm}}} 
                   & 13B & 53.4 & 30.2 & 53.4 & 30.2 & 35.6 & 36.1 & 35.9 \\
                   & 70B & 62.7 & 42.1 & 62.6 & 42.0 & 38.2 & 37.2 & 37.6 \\
            \midrule
            \multirow{2}{*}{\makecell{Xwin-LM\\~\cite{teamxwin}}} & 7B & 45.5 & 16.1 & 45.5 & 16.1 & 33.8 & 34.3 & 34.1 \\
                 & 13B & 53.9 & 27.2 & 53.9 & 27.1 & 36.6 & 36.5 & 35.9 \\
            \midrule
            \multirow{2}{*}{\makecell{Alpaca\\~\cite{taori2023alpaca}}} & 7B & 40.8 & 13.7 & 40.7 & 13.7 & 34.7 & 35.3 & 35.4 \\
                   & 13B & 39.5 & 14.8 & 37.8 & 13.7 & 38.0 & 37.4 & 38.7 \\
            \midrule
            \multirow{3}{*}{\makecell{Vicuna\\~\cite{chiang2023vicuna}}} & 7B & 48.8 & 25.0 & 48.8 & 24.9 & 33.6 & 34.2 & 34.1 \\
                   & 13B & 54.5 & 33.5 & 54.5 & 33.4 & 35.7 & 35.8 & 35.5 \\
                   & 33B & 57.1 & 36.3 & 57.0 & 36.3 & 36.0 & 36.1 & 36.0 \\
            \bottomrule         
        \end{tabular}
    \label{tab:MMLU-0-shot}
\end{table*}

\begin{table*}[t]
    \centering
    \caption{MCQA evaluation on 0-shot HellaSwag~\cite{zellers2019hellaswag}.
    } \vspace{-3mm}
        \begin{tabular}{ccccccccc}
            \toprule
            Model & Size & Choices & \makecell{Choices\\(Circular)} & Vocab & \makecell{Vocab\\(Circular)} & Alignment & \makecell{Normalized \\Alignment} & PPL \\
             \midrule
             \multirow{2}{*}{\makecell{MPT\\~\cite{team2023introducing, mosaicml2023introducing}}} & 7B & 28.1 & 0.8 & 28.1 & 0.6 & 57.1 & 76.3 & 75.8 \\
                & 30B & 32.8 & 3.3 & 32.8 & 2.9 & 60.4 & 79.9 & 79.3 \\
             \midrule
             \makecell{MPT-Chat\\~\cite{mosaicml2023introducing}} & 30B & 46.9 & 14.9 & 46.9 & 15.2 & 61.5 & 80.1 & 79.7  \\
             \midrule
             \multirow{2}{*}{\makecell{Falcon\\~\cite{almazrouei2023falcon}}} & 7B & 25.1 & 0.0 & 24.1 & 0.0 & 57.7 & 76.3 & 75.9 \\
                & 40B & 48.7 & 16.7& 48.3 & 16.5& 64.0 & 82.8& 82.6\\
             \midrule
             \multirow{4}{*}{\makecell{LLaMA-1\\~\cite{touvron2023llama}}} & 7B & 29.1 & 1.1 & 28.9 & 1.1 & 56.9 & 76.2 & 74.8 \\
                & 13B & 33.1 & 3.2 & 33.1 & 3.2 & 59.9 & 79.1 & 78.1 \\
                & 30B & 44.3 & 14.7 & 44.4 & 14.8 & 63.3 & 82.6 & 81.2 \\
                & 65B & 46.2 & 12.7 & 45.8 & 12.7 & 64.5 & 84.1 & 82.9 \\
            \midrule
            \multirow{3}{*}{\makecell{LLaMA-2\\~\cite{touvron2023llama2}}} & 7B & 32.1 & 1.2 & 30.0 & 1.2 & 57.2 & 76.0 & 75.0 \\
                    & 13B & 49.7 & 17.2 & 48.6 & 16.6 & 60.1 & 79.4 & 78.2 \\
                    & 70B & 62.4 & 31.1 & 62.1 & 31.0 & 64.8 & 83.8 & 82.7 \\
            \midrule
            \multirow{3}{*}{\makecell{LLaMA-2-Chat\\~\cite{touvron2023llama2}}} & 7B & 50.5 & 16.9 & 42.1 & 13.7 & 57.7 & 75.4 & 75.1 \\
                         & 13B & 63.5 & 34.9 & 63.5 & 34.8 & 60.7 & 79.7 & 78.7 \\
                         & 70B & 75.2 & 55.3 & 75.2 & 55.3 & 63.8 & 82.2 & 81.3 \\
            \midrule
            \multirow{2}{*}{\makecell{WizardLM\\~\cite{xu2023wizardlm}}} & 13B & 66.5 & 44.4 & 66.5 & 44.4 & 61.5 & 79.8 & 79.1 \\
                   & 70B & 70.9 & 46.7 & 70.9 & 46.7 & 64.8 & 82.1 & 81.5 \\
            \midrule
            \multirow{2}{*}{\makecell{Xwin-LM\\~\cite{teamxwin}}} & 7B & 39.6 & 6.0 & 39.6 & 6.0 & 58.8 & 76.8 & 76.0 \\
                 & 13B & 56.1 & 22.7 & 56.1 & 22.6 & 62.1 & 80.9 & 79.8 \\
            \midrule
            \multirow{2}{*}{\makecell{Alpaca\\~\cite{taori2023alpaca}}} & 7B & 37.3 & 7.7 & 37.2 & 7.7 & 59.1 & 75.6 & 75.3 \\
                   & 13B & 38.9 & 11.3 & 38.7 & 11.3 & 60.6 & 78.3 & 77.8 \\
            \midrule
            \multirow{3}{*}{\makecell{Vicuna\\~\cite{chiang2023vicuna}}} & 7B & 55.8 & 30.2 & 55.9 & 30.2 & 56.4 & 73.8 & 73.0 \\
                   & 13B & 61.2 & 32.4 & 61.2 & 32.5 & 59.6 & 77.5 & 76.8 \\
                   & 33B & 65.1 & 35.1 & 65.1 & 35.2 & 61.9 & 80.4 & 79.6 \\
            \bottomrule        
        \end{tabular}
    %}

    \label{tab:HellaSwag-0-shot}
\end{table*}

\begin{table*}[h]
    \centering
        \caption{MCQA evaluation on 0-shot ARC-Challenge~\cite{clark2018think}.} \vspace{-3mm}
    %\resizebox{\textwidth/2}{!}{%
        \begin{tabular}{ccccccccc}
            \toprule
            Model & Size & Choices & \makecell{Choices\\(Circular)} & Vocab & \makecell{Vocab\\(Circular)} & Alignment & \makecell{Normalized \\Alignment} & PPL \\
             \midrule
             \multirow{2}{*}{\makecell{MPT\\~\cite{team2023introducing, mosaicml2023introducing}}} & 7B & 31.8 & 0.3 & 30.4 & 0.3 & 40.0 & 41.8 & 39.8 \\
                & 30B & 51.0 & 22.4 & 50.3 & 21.0 & 46.9 & 49.4 & 49.6 \\
             \midrule
             \makecell{MPT-Chat\\~\cite{mosaicml2023introducing}} & 30B & 66.0 & 43.2 & 65.5 & 42.7 & 50.0 & 51.0 & 50.8 \\
             \midrule
             \multirow{2}{*}{\makecell{Falcon\\~\cite{almazrouei2023falcon}}} & 7B & 27.5 & 0.0 & 27.0 & 0.0 & 40.3 & 43.6 & 42.9 \\
                & 40B & 64.2 & 41.7 & 64.2 & 41.6 & 50.3 & 54.5 & 52.6 \\
             \midrule
             \multirow{4}{*}{\makecell{LLaMA-1\\~\cite{touvron2023llama}}} & 7B & 38.9 & 1.9 & 38.1 & 2.0 & 41.9 & 44.6 & 41.4 \\
                & 13B & 52.4 & 21.3 & 51.5 & 20.6 & 46.4 & 47.8 & 47.2 \\
                & 30B & 69.4 & 48.4 & 69.2 & 48.3 & 52.9 & 52.9 & 50.8 \\
                & 65B & 74.3 & 52.6 & 74.3 & 52.6 & 52.8 & 55.7 & 52.5 \\
            \midrule
            \multirow{3}{*}{\makecell{LLaMA-2\\~\cite{touvron2023llama2}}} & 7B & 45.9 & 16.4 & 44.8 & 15.3 & 43.4 & 46.2 & 43.0 \\
                    & 13B & 63.4 & 30.5 & 62.5 & 29.3 & 48.4 & 49.2 & 47.7 \\
                    & 70B & 81.7 & 67.1 & 81.7 & 67.1 & 54.4 & 57.3 & 53.8 \\
            \midrule
            \multirow{3}{*}{\makecell{LLaMA-2-Chat\\~\cite{touvron2023llama2}}} & 7B & 55.8 & 23.0 & 55.7 & 23.0 & 44.1 & 44.5 & 42.7 \\
                         & 13B & 64.3 & 38.1 & 64.2 & 37.9 & 46.2 & 50.3 & 46.9 \\
                         & 70B & 78.2 & 64.2 & 78.1 & 64.2 & 53.0 & 54.4 & 51.1\\
            \midrule
            \multirow{2}{*}{\makecell{WizardLM\\~\cite{xu2023wizardlm}}}
                   & 13B & 65.5 & 43.4 & 65.5 & 43.3 & 47.0 & 50.1 & 47.1 \\
                   & 70B & 80.6 & 65.7 & 80.6 & 65.6 & 53.2 & 56.7 & 53.5 \\
            \midrule
            \multirow{2}{*}{\makecell{Xwin-LM\\~\cite{teamxwin}}} & 7B & 54.9 & 23.5 & 54.9 & 23.5 & 45.9 & 47.6 & 44.7 \\
                 & 13B & 67.4 & 36.8 & 67.2 & 35.9 & 52.0 & 53.0 & 51.6 \\
            \midrule
            \multirow{2}{*}{\makecell{Alpaca\\~\cite{taori2023alpaca}}} & 7B & 48.5 & 17.2 & 48.2 & 17.0 & 44.9 & 46.5 & 45.1 \\
                   & 13B & 47.4 & 22.8 & 47.2 & 22.8 & 46.1 & 48.4 & 46.8 \\
            \midrule
            \multirow{3}{*}{\makecell{Vicuna\\~\cite{chiang2023vicuna}}} & 7B & 60.4 & 33.9 & 60.6 & 33.9 & 43.3 & 45.7 & 42.1 \\
                   & 13B & 70.0 & 48.0 & 70.0 & 48.1 & 47.8 & 50.7 & 48.8 \\
                   & 33B & 71.2 & 49.9 & 71.1 & 49.6 & 50.2 & 51.6 & 50.7 \\
            \bottomrule        
        \end{tabular}
    %}
    \label{tab:ARC-c-0-shot}
\end{table*}

\begin{table*}[h]
    \centering
    \caption{MCQA evaluation on 0-shot ARC-Easy~\cite{clark2018think}.} \vspace{-3mm}
        \begin{tabular}{ccccccccc}
            \toprule
            Model & Size & Choices & \makecell{Choices\\(Circular)} & Vocab & \makecell{Vocab\\(Circular)} & Alignment & \makecell{Normalized \\Alignment} & PPL \\
             \midrule
             \multirow{2}{*}{\makecell{MPT\\~\cite{team2023introducing, mosaicml2023introducing}}} & 7B  & 36.4 & 0.8 & 35.9 & 0.6 & 74.7 & 70.2 & 69.3 \\
                & 30B & 70.0 & 42.2 & 69.9 & 40.7 & 78.7 & 76.3 & 74.0 \\
             \midrule
             \makecell{MPT-Chat\\~\cite{mosaicml2023introducing}} & 30B & 79.4 & 62.6 & 79.5 & 62.7 & 80.2 & 77.4 & 75.7 \\
             \midrule
             \multirow{2}{*}{\makecell{Falcon\\~\cite{almazrouei2023falcon}}} & 7B & 29.2 & 0.0 & 29.0 & 0.0 & 74.7 & 70.8 & 69.9 \\
                & 40B & 80.0 & 61.7 & 80.0 & 61.6 & 81.9 & 79.3 & 77.7 \\
             \midrule
             \multirow{4}{*}{\makecell{LLaMA-1\\~\cite{touvron2023llama}}} & 7B & 44.7 & 5.2 & 43.8 & 4.6 & 75.3 & 72.9 & 67.7 \\
                & 13B & 67.6 & 36.7 & 66.7 & 35.4 & 77.4 & 74.8 & 69.2 \\
                & 30B & 84.6 & 69.6 & 84.7 & 69.6 & 80.4 & 79.0 & 73.5 \\
                & 65B & 87.7 & 73.0 & 87.7 & 72.9 & 81.3 & 79.8 & 75.1 \\
            \midrule
            \multirow{3}{*}{\makecell{LLaMA-2\\~\cite{touvron2023llama2}}} & 7B & 59.0 & 26.8 & 58.2 & 25.8 & 76.3 & 74.6 & 68.9 \\
                    & 13B & 77.4 & 53.7 & 76.6 & 52.1 & 79.4 & 77.5 & 71.9 \\
                    & 70B & 92.8 & 84.5 & 92.8 & 84.4 & 82.7 & 81.0 & 75.9 \\
            \midrule
            \multirow{3}{*}{\makecell{LLaMA-2-Chat\\~\cite{touvron2023llama2}}} & 7B & 70.9 & 44.9 & 70.8 & 44.9 & 74.4 & 69.7 & 66.9 \\
                         & 13B & 79.9 & 62.0 & 79.9 & 62.0 & 77.5 & 73.7 & 70.5 \\
                         & 70B & 92.8 & 84.5 & 92.8 & 84.4 & 82.7 & 81.0 & 75.9 \\
            \midrule
            \multirow{2}{*}{\makecell{WizardLM\\~\cite{xu2023wizardlm}}} 
                   & 13B & 81.9 & 67.2 & 81.9 & 67.2 & 78.5 & 74.2 & 69.7 \\
                   & 70B & 92.1 & 83.9 & 92.1 & 83.9 & 81.9 & 77.9 & 73.6 \\
            \midrule
            \multirow{2}{*}{\makecell{Xwin-LM\\~\cite{teamxwin}}} & 7B & 69.7 & 41.2 & 69.7 & 41.1 & 76.9 & 74.3 & 68.6 \\
                 & 13B & 80.5 & 60.7 & 80.3 & 60.1 & 81.2 & 78.0 & 73.7 \\
            \midrule
            \multirow{2}{*}{\makecell{Alpaca\\~\cite{taori2023alpaca}}} & 7B & 67.7 & 36.2 & 67.2 & 35.6 & 74.8 & 70.6 & 66.9 \\
                   & 13B & 67.2 & 41.3 & 67.0 & 40.9 & 76.4 & 71.7 & 68.0 \\
            \midrule
            \multirow{3}{*}{\makecell{Vicuna\\~\cite{chiang2023vicuna}}} & 7B & 75.9 & 55.0 & 75.9 & 55.0 & 75.6 & 71.3 & 66.6 \\
                   & 13B & 83.8 & 71.4 & 83.8 & 71.4 & 78.7 & 74.8 & 70.5 \\
                   & 33B & 85.9 & 70.7 & 85.6 & 69.6 & 78.4 & 74.7 & 70.2 \\
            \bottomrule        
        \end{tabular}
    %}
    \label{tab:ARC-e-0-shot}
\end{table*}

\begin{table*}[h]
    \centering
    \caption{MCQA evaluation on 0-shot BoolQ~\cite{clark2019boolq}.} \vspace{-3mm}
        \begin{tabular}{ccccccccc}
            \toprule
            Model & Size & Choices & \makecell{Choices\\(Circular)} & Vocab & \makecell{Vocab\\(Circular)} & Alignment & \makecell{Normalized \\Alignment} & PPL \\
            \midrule
            \multirow{2}{*}{\makecell{MPT\\~\cite{team2023introducing, mosaicml2023introducing}}} & 7B & 62.1 & 0.6 & 57.2 & 0.2 & 74.3 & 67.9 & 74.3 \\
                & 30B & 69.1 & 47.9 & 30.6 & 19.8 & 74.1 & 69.2 & 74.1 \\
            \midrule
            \makecell{MPT-Chat\\~\cite{mosaicml2023introducing}} & 30B & 65.6 & 40.1 & 61.6 & 32.0 & 80.1 & 81.6 & 80.1 \\
            \midrule
            \multirow{2}{*}{\makecell{Falcon\\~\cite{almazrouei2023falcon}}} & 7B & 57.3 & 28.1 & 45.4 & 20.5 & 73.6 & 64.5 & 73.6 \\
                & 40B & 67.7 & 22.6 & 67.5 & 22.2 & 82.0 & 71.4 & 82.0 \\
            \midrule
            \multirow{4}{*}{\makecell{LLaMA-1\\~\cite{touvron2023llama}}} & 7B & 59.9 & 34.1 & 50.9 & 28.0 & 75.0 & 66.6 & 75.0 \\
                & 13B & 64.3 & 22.5 & 53.1 & 14.2 & 78.0 & 65.6 & 78.0 \\
                & 30B & 78.5 & 70.8 & 77.2 & 68.6 & 82.7 & 70.2 & 82.7 \\
                & 65B & 79.9 & 67.4 & 71.4 & 59.1 & 84.8 & 74.6 & 84.8 \\
            \midrule
            \multirow{3}{*}{\makecell{LLaMA-2\\~\cite{touvron2023llama2}}} & 7B & 63.2 & 46.7 & 24.7 & 16.3 & 77.7 & 64.9 & 77.7 \\
                    & 13B & 66.9 & 10.1 & 49.8 & 4.7 & 80.6 & 66.2 & 80.6 \\
                    & 70B & 85.0 & 76.5 & 85.0 & 76.5 & 83.7 & 70.9 & 83.7 \\
            \midrule
            \multirow{3}{*}{\makecell{LLaMA-2-Chat\\~\cite{touvron2023llama2}}} & 7B & 63.6 & 21.2 & 63.6 & 21.2 & 80.7 & 79.4 & 80.7 \\
                         & 13B & 71.9 & 39.3 & 71.9 & 39.2 & 81.7 & 83.1 & 81.7 \\
                         & 70B & 75.6 & 57.8 & 75.6 & 57.8 & 86.7 & 82.8 & 86.7 \\
            \midrule
            \multirow{2}{*}{\makecell{WizardLM\\~\cite{xu2023wizardlm}}} 
                   & 13B & 76.5 & 51.7 & 76.5 & 51.7 & 84.0 & 77.7 & 84.0 \\
                   & 70B & 89.0 & 84.5 & 89.0 & 84.5 & 86.6 & 77.5 & 86.6 \\
            \midrule
            \multirow{2}{*}{\makecell{Xwin-LM\\~\cite{teamxwin}}} & 7B & 64.2 & 40.2 & 63.8 & 39.8 & 79.3 & 64.8 & 79.3 \\
                 & 13B & 68.5 & 15.1 & 68.3 & 14.9 & 83.5 & 72.0 & 83.5 \\
            \midrule
            \multirow{2}{*}{\makecell{Alpaca\\~\cite{taori2023alpaca}}} & 7B & 64.2 & 38.8 & 64.1 & 38.7 & 77.5 & 79.4 & 77.5 \\
                   & 13B & 68.0 & 61.2 & 68.0 & 61.0 & 78.8 & 74.3 & 78.8 \\
            \midrule
            \multirow{3}{*}{\makecell{Vicuna\\~\cite{chiang2023vicuna}}} & 7B & 67.7 & 54.0 & 67.7 & 54.0 & 80.9 & 68.0 & 80.9 \\
                   & 13B & 82.1 & 62.6 & 82.1 & 62.6 & 85.2 & 80.0 & 85.2 \\
                   & 33B & 80.2 & 74.0 & 27.9 & 23.0 & 83.9 & 76.1 & 83.9 \\
            \bottomrule        
        \end{tabular}
    %}
    \label{tab:BoolQ-0-shot}
\end{table*}

\begin{table*}[h]
    \centering
    \caption{MCQA evaluation on 0-shot SIQA~\cite{sap2019socialiqa}.} \vspace{-3mm}
        \begin{tabular}{ccccccccc}
            \toprule
            Model & Size & Choices & \makecell{Choices\\(Circular)} & Vocab & \makecell{Vocab\\(Circular)} & Alignment & \makecell{Normalized \\Alignment} & PPL \\
             \midrule
             \multirow{2}{*}{\makecell{MPT\\~\cite{team2023introducing, mosaicml2023introducing}}} & 7B & 37.7 & 2.4 & 38.6 & 2.0 & 45.6 & 48.2 & 45.9 \\
                & 30B & 58.5 & 34.9 & 57.3 & 31.9 & 46.0 & 48.9 & 45.6 \\
             \midrule
            \makecell{MPT-Chat\\~\cite{mosaicml2023introducing}} & 30B & 63.4 & 42.5 & 63.2 & 41.6 & 48.8 & 49.6 & 48.3 \\
            \midrule
             \multirow{2}{*}{\makecell{Falcon\\~\cite{almazrouei2023falcon}}} & 7B & 36.5 & 1.2 & 36.5 & 1.2 & 45.4 & 48.7 & 47.0 \\
                & 40B & 65.6 & 46.8 & 65.6 & 46.7 & 49.2 & 51.0 & 47.8 \\
             \midrule
             \multirow{4}{*}{\makecell{LLaMA-1\\~\cite{touvron2023llama}}} & 7B & 44.6 & 11.5 & 44.5 & 11.6 & 44.8 & 47.0 & 44.3 \\
                & 13B & 53.0 & 14.8 & 53.2 & 14.8 & 45.0 & 48.0 & 44.3 \\
                & 30B & 66.1 & 49.0 & 66.0 & 48.8 & 45.7 & 49.4 & 45.2 \\
                & 65B & 67.2 & 67.1 & 49.7 & 49.8 & 47.5 & 50.1 & 46.3 \\
            \midrule
            \multirow{3}{*}{\makecell{LLaMA-2\\~\cite{touvron2023llama2}}} & 7B & 51.4 & 21.2 & 51.0 & 20.6 & 43.5 & 47.3 & 44.2 \\
                    & 13B & 58.5 & 36.4 & 58.3 & 36.1 & 44.7 & 48.4 & 44.8 \\
                    & 70B & 70.9 & 55.3 & 70.8 & 55.2 & 46.3 & 49.4 & 45.3 \\
            \midrule
            \multirow{3}{*}{\makecell{LLaMA-2-Chat\\~\cite{touvron2023llama2}}} & 7B & 56.9 & 31.8 & 56.9 & 31.9 & 46.1 & 48.6 & 45.1 \\
                         & 13B & 62.6 & 41.6 & 62.7 & 41.5 & 49.2 & 50.3 & 46.0 \\
                         & 70B & 67.2 & 46.7 & 67.2 & 46.7 & 49.2 & 50.6 & 47.4 \\
            \midrule
            \multirow{2}{*}{\makecell{WizardLM\\~\cite{xu2023wizardlm}}} 
                   & 13B & 66.7 & 51.5 & 66.7 & 51.6 & 49.3 & 49.2 & 46.7 \\
                   & 70B & 73.8 & 59.1 & 73.8 & 59.1 & 49.8 & 49.8 & 47.9 \\
            \midrule
            \multirow{2}{*}{\makecell{Xwin-LM\\~\cite{teamxwin}}} & 7B & 56.9 & 28.1 & 56.9 & 28.0 & 47.5 & 49.3 & 45.5 \\
                 & 13B & 56.8 & 34.3 & 56.7 & 34.1 & 48.5 & 48.4 & 45.4 \\
            \midrule
            \multirow{2}{*}{\makecell{Alpaca\\~\cite{taori2023alpaca}}} & 7B & 50.7 & 23.8 & 50.7 & 23.8 & 48.3 & 49.2 & 45.3 \\
                   & 13B & 56.8 & 34.3 & 56.7 & 34.1 & 48.5 & 48.4 & 45.4 \\
            \midrule
            \multirow{3}{*}{\makecell{Vicuna\\~\cite{chiang2023vicuna}}} & 7B & 63.3 & 42.5 & 63.3 & 42.3 & 45.8 & 47.6 & 44.3 \\
                   & 13B & 67.2 & 50.3 & 67.1 & 50.5 & 46.6 & 47.3 & 44.6 \\
                   & 33B & 63.2 & 42.2 & 62.5 & 41.1 & 46.6 & 48.2 & 46.0 \\
            \bottomrule        
        \end{tabular}
    %}
    \label{tab:SIQA-0-shot}
\end{table*}

\begin{table*}[t]
    \centering
    \caption{MCQA evaluation on 0-shot PIQA~\cite{bisk2020piqa}.}\vspace{-3mm}
        \begin{tabular}{ccccccccc}
            \toprule
            Model & Size & Choices & \makecell{Choices\\(Circular)} & Vocab & \makecell{Vocab\\(Circular)} & Alignment & \makecell{Normalized \\Alignment} & PPL \\
             \midrule
             \multirow{2}{*}{\makecell{MPT\\~\cite{team2023introducing, mosaicml2023introducing}}} & 7B & 55.3 & 16.0 & 54.2 & 15.6 & 79.3 & 80.7 & 80.1 \\
                & 30B & 64.9 & 39.8 & 64.7 & 39.6 & 80.1 & 81.1 & 81.1 \\
             \midrule
            \makecell{MPT-Chat\\~\cite{mosaicml2023introducing}} & 30B & 65.1 & 34.4 & 62.6 & 33.4 & 79.2 & 81.0 & 80.8 \\
            \midrule
             \multirow{2}{*}{\makecell{Falcon\\~\cite{almazrouei2023falcon}}} & 7B & 50.4 & 10.2 & 49.8 & 10.0 & 79.3 & 80.4 & 80.0 \\
                & 40B & 69.7 & 44.6 & 69.5 & 44.4 & 82.3 & 83.1 & 82.9 \\
             \midrule
             \multirow{4}{*}{\makecell{LLaMA-1\\~\cite{touvron2023llama}}} & 7B & 53.8 & 10.9 & 53.5 & 10.9 & 78.7 & 79.2 & 78.8 \\
                & 13B & 64.1 & 32.3 & 63.2 & 32.0 & 79.2 & 80.2 & 79.6 \\
                & 30B & 73.3 & 52.1 & 73.3 & 52.0 & 81.0 & 82.2 & 81.3 \\
                & 65B & 66.2 & 33.9 & 66.2 & 33.9 & 81.3 & 82.3 & 82.3 \\
            \midrule
            \multirow{3}{*}{\makecell{LLaMA-2\\~\cite{touvron2023llama2}}} & 7B & 59.4 & 21.4 & 58.8 & 20.8 & 78.1 & 79.1 & 78.6 \\
                    & 13B & 71.7 & 46.0 & 71.4 & 45.9 & 79.1 & 80.5 & 80.4 \\
                    & 70B & 76.1 & 56.8 & 76.1 & 56.8 & 82.2 & 82.7 & 82.8 \\
            \midrule
            \multirow{3}{*}{\makecell{LLaMA-2-Chat\\~\cite{touvron2023llama2}}} & 7B & 65.0 & 35.1 & 64.4 & 34.8 & 76.8 & 76.7 & 76.7 \\
                         & 13B & 70.7 & 45.1 & 70.1 & 44.5 & 77.6 & 79.1 & 78.3 \\
                         & 70B & 78.0 & 58.9 & 76.9 & 57.4 & 80.6 & 80.8 & 80.7 \\
            \midrule
            \multirow{2}{*}{\makecell{WizardLM\\~\cite{xu2023wizardlm}}}
                   & 13B & 75.7 & 56.4 & 75.7 & 56.4 & 79.0 & 79.4 & 79.7 \\
                   & 70B & 82.4 & 70.3 & 82.2 & 70.0 & 81.1 & 80.7 & 81.0 \\
            \midrule
            \multirow{2}{*}{\makecell{Xwin-LM\\~\cite{teamxwin}}} & 7B & 67.5 & 40.0 & 66.8 & 39.5 & 78.0 & 78.5 & 78.7 \\
                 & 13B & 69.2 & 41.9 & 68.8 & 41.6 & 79.2 & 80.8 & 80.0 \\
            \midrule
            \multirow{2}{*}{\makecell{Alpaca\\~\cite{taori2023alpaca}}} & 7B & 64.5 & 35.5 & 64.5 & 35.5 & 77.8 & 78.0 & 77.7 \\
                   & 13B & 63.1 & 32.4 & 63.0 & 32.0 & 78.1 & 78.6 & 77.4 \\
            \midrule
            \multirow{3}{*}{\makecell{Vicuna\\~\cite{chiang2023vicuna}}} & 7B & 72.2 & 48.3 & 72.2 & 48.3 & 77.3 & 78.0 & 78.3 \\
                   & 13B & 75.3 & 56.4 & 75.3 & 56.4 & 78.9 & 79.2 & 79.1 \\
                   & 33B & 73.8 & 53.3 & 70.0 & 50.7 & 79.2 & 79.4 & 79.5 \\
            \bottomrule        
        \end{tabular}
    %}
    \label{tab:PIQA-0-shot}
\end{table*}

\begin{table*}[t]
    \centering
    \caption{MCQA evaluation on 0-shot AGIEval (English only)~\cite{zhong2023agieval}.} \vspace{-3mm}
        \begin{tabular}{ccccccccc}
            \toprule
            Model & Size & Choices & \makecell{Choices\\(Circular)} & Vocab & \makecell{Vocab\\(Circular)} & Alignment & \makecell{Normalized \\Alignment} & PPL \\
             \midrule
             \multirow{2}{*}{\makecell{MPT\\~\cite{team2023introducing, mosaicml2023introducing}}} & 7B & 23.5 & 0.0 & 21.8 & 0.0 & 25.7 & 26.9 & 26.9 \\
                & 30B & 24.0 & 0.7 & 24.0 & 0.6 & 27.7 & 28.9 & 28.9 \\
             \midrule
            \makecell{MPT-Chat\\~\cite{mosaicml2023introducing}} & 30B & 29.1 & 3.2 & 29.0 & 3.2 & 28.2 & 29.7 & 29.1 \\
            \midrule

             \multirow{2}{*}{\makecell{Falcon\\~\cite{almazrouei2023falcon}}} & 7B & 22.3 & 0.0 & 20.9 & 0.0 & 25.2 & 27.2 & 26.7 \\
                & 40B & 28.8 & 3.0 & 28.2 & 3.0 & 28.5 & 30.8 & 31.4 \\
             \midrule
             \multirow{4}{*}{\makecell{LLaMA-1\\~\cite{touvron2023llama}}} & 7B & 22.0 & 0.0 & 21.9 & 0.0 & 25.8 & 27.7 & 26.9 \\
                & 13B & 26.6 & 1.1 & 26.4 & 1.1 & 28.0 & 29.5 & 29.7 \\
                & 30B & 33.1 & 7.8 & 32.8 & 7.7 & 30.0 & 31.2 & 31.0 \\
                & 65B & 38.7 & 10.1 & 37.9 & 10.0 & 30.4 & 32.7 & 32.3 \\
            \midrule
            \multirow{3}{*}{\makecell{LLaMA-2\\~\cite{touvron2023llama2}}} & 7B & 24.4 & 0.6 & 22.1 & 0.5 & 26.4 & 29.1 & 28.7 \\
                    & 13B & 33.9 & 8.0 & 34.0 & 8.0 & 28.0 & 30.3 & 30.1 \\
                    & 70B & 50.0 & 26.2 & 49.9 & 26.2 & 31.7 & 34.1 & 34.6 \\
            \midrule
            \multirow{3}{*}{\makecell{LLaMA-2-Chat\\~\cite{touvron2023llama2}}} & 7B & 28.3 & 2.9 & 17.3 & 1.8 & 25.9 & 27.1 & 26.6 \\
                         & 13B & 35.9 & 9.7 & 35.5 & 9.7 & 26.5 & 29.2 & 28.8 \\
                         & 70B & 45.9 & 17.6 & 45.8 & 17.6 & 29.7 & 31.1 & 32.1 \\
            \midrule
            \multirow{2}{*}{\makecell{WizardLM\\~\cite{xu2023wizardlm}}} 
                   & 13B & 37.6 & 8.6 & 37.5 & 8.6 & 27.1 & 29.6 & 30.0 \\
                   & 70B & 48.2 & 22.5 & 47.6 & 21.9 & 30.6 & 32.6 & 32.7 \\
            \midrule
            \multirow{2}{*}{\makecell{Xwin-LM\\~\cite{teamxwin}}} & 7B & 30.9 & 3.1 & 29.5 & 2.9 & 28.1 & 29.9 & 29.5 \\
                 & 13B & 35.4 & 10.4 & 35.4 & 10.4 & 29.0 & 31.0 & 30.8 \\
            \midrule
            \multirow{2}{*}{\makecell{Alpaca\\~\cite{taori2023alpaca}}} & 7B & 24.1 & 0.7 & 23.9 & 0.7 & 26.8 & 28.9 & 27.7 \\
                   & 13B & 27.8 & 3.1 & 27.2 & 2.9 & 26.5 & 28.7 & 28.6 \\
            \midrule
            \multirow{3}{*}{\makecell{Vicuna\\~\cite{chiang2023vicuna}}} & 7B & 35.5 & 10.9 & 33.4 & 10.3 & 26.6 & 28.1 & 28.2 \\
                   & 13B & 38.6 & 14.7 & 37.1 & 14.5 & 27.5 & 29.6 & 28.3 \\
                   & 33B & 39.9 & 14.4 & 39.7 & 14.3 & 28.6 & 30.3 & 31.1 \\
            \bottomrule        
        \end{tabular}
    %}
    \label{tab:AGIEval-0-shot}
\end{table*}

\begin{table*}[t]
    \centering
    \caption{MCQA evaluation on 0-shot OpenBookQA (with fact)~\cite{mihaylov2018can}.} \vspace{-3mm}
        \begin{tabular}{ccccccccc}
            \toprule
            Model & Size & Choices & \makecell{Choices\\(Circular)} & Vocab & \makecell{Vocab\\(Circular)} & Alignment & \makecell{Normalized \\Alignment} & PPL \\
             \midrule
             \multirow{2}{*}{\makecell{MPT\\~\cite{team2023introducing, mosaicml2023introducing}}} & 7B & 39.0 & 1.6 & 38.8 & 1.4 & 41.6 & 52.4 & 52.8 \\
                & 30B & 70.4 & 43.0 & 69.6 & 41.8 & 45.2 & 53.8 & 55.4 \\
             \midrule
            \makecell{MPT-Chat\\~\cite{mosaicml2023introducing}} & 30B & 77.8 & 58.6 & 77.4 & 58.0 & 47.0 & 54.4 & 55.2 \\
            \midrule
             \multirow{2}{*}{\makecell{Falcon\\~\cite{almazrouei2023falcon}}} & 7B & 28.2 & 0.0 & 28.0 & 0.0 & 43.2 & 53.0 & 54.2 \\
                & 40B & 77.2 & 61.0 & 77.2 & 61.0 & 48.0 & 55.6 & 58.4 \\
             \midrule
             \multirow{4}{*}{\makecell{LLaMA-1\\~\cite{touvron2023llama}}} & 7B & 49.2 & 17.8 & 48.2 & 16.8 & 44.8 & 52.4 & 54.0 \\
                & 13B & 64.8 & 32.8 & 64.6 & 32.4 & 44.4 & 53.4 & 53.0 \\
                & 30B & 81.6 & 66.6 & 81.6 & 66.6 & 44.0 & 54.6 & 55.4 \\
                & 65B & 82.8 & 67.6 & 82.4 & 67.0 & 46.0 & 54.6 & 57.2 \\
            \midrule
            \multirow{3}{*}{\makecell{LLaMA-2\\~\cite{touvron2023llama2}}} & 7B & 62.6 & 28.8 & 61.8 & 27.8 & 44.6 & 52.6 & 53.4 \\
                    & 13B & 72.0 & 50.2 & 71.8 & 48.8 & 44.4 & 54.2 & 54.0 \\
                    & 70B & 88.4 & 79.2 & 86.6 & 77.2 & 47.6 & 56.6 & 58.0 \\
            \midrule
            \multirow{3}{*}{\makecell{LLaMA-2-Chat\\~\cite{touvron2023llama2}}} & 7B & 73.8 & 52.4 & 73.8 & 52.4 & 48.4 & 54.8 & 55.0 \\
                         & 13B & 80.0 & 63.4 & 80.0 & 63.4 & 48.6 & 56.4 & 59.0 \\
                         & 70B & 86.0 & 76.4 & 86.0 & 76.4 & 48.6 & 58.6 & 59.0 \\
            \midrule
            \multirow{2}{*}{\makecell{WizardLM\\~\cite{xu2023wizardlm}}} 
                   & 13B & 79.4 & 65.0 & 79.4 & 65.2 & 48.0 & 56.8 & 57.0 \\
                   & 70B & 86.6 & 75.6 & 86.6 & 75.6 & 48.8 & 57.8 & 57.0 \\
            \midrule
            \multirow{2}{*}{\makecell{Xwin-LM\\~\cite{teamxwin}}} & 7B & 67.2 & 37.2 & 67.0 & 36.8 & 46.0 & 56.2 & 54.4 \\
                 & 13B & 78.0 & 53.2 & 78.0 & 53.0 & 44.8 & 55.4 & 55.0 \\
            \midrule
            \multirow{2}{*}{\makecell{Alpaca\\~\cite{taori2023alpaca}}} & 7B & 73.8 & 48.8 & 73.4 & 47.6 & 47.4 & 54.4 & 56.4 \\
                   & 13B & 73.0 & 52.4 & 72.6 & 51.8 & 47.4 & 55.6 & 55.8 \\
            \midrule
            \multirow{3}{*}{\makecell{Vicuna\\~\cite{chiang2023vicuna}}} & 7B & 79.0 & 79.2 & 63.0 & 63.2 & 47.2 & 54.0 & 54.8 \\
                   & 13B & 79.4 & 65.0 & 79.4 & 65.2 & 48.0 & 56.8 & 57.0 \\
                   & 33B & 86.6 & 75.6 & 86.6 & 75.6 & 48.8 & 57.8 & 57.0 \\
            \bottomrule        
        \end{tabular}
    %}

    \label{tab:OpenBookQAFact-0-shot}
\end{table*}

\begin{table*}[t]
    \centering
    \caption{MCQA evaluation on 0-shot CommonSenseQA~\cite{talmor2018commonsenseqa}.} \vspace{-3mm}
        \begin{tabular}{ccccccccc}
            \toprule
            Model & Size & Choices & \makecell{Choices\\(Circular)} & Vocab & \makecell{Vocab\\(Circular)} & Alignment & \makecell{Normalized \\Alignment} & PPL \\
             \midrule
             \multirow{2}{*}{\makecell{MPT\\~\cite{team2023introducing, mosaicml2023introducing}}} & 7B & 22.9 & 0.3 & 22.8 & 0.2 & 56.5 & 49.1 & 47.2 \\
                & 30B & 40.5 & 10.9 & 40.1 & 10.3 & 61.5 & 55.0 & 51.4 \\
             \midrule
            \makecell{MPT-Chat\\~\cite{mosaicml2023introducing}} & 30B & 64.1 & 39.0 & 64.1 & 38.9 & 64.0 & 54.6 & 52.5 \\
            \midrule
             \multirow{2}{*}{\makecell{Falcon\\~\cite{almazrouei2023falcon}}} & 7B & 20.7 & 0.0 & 20.9 & 0.0 & 57.9 & 50.0 & 50.0 \\
                & 40B & 62.9 & 33.5 & 62.8 & 33.4 & 62.5 & 54.9 & 52.2 \\
             \midrule
             \multirow{4}{*}{\makecell{LLaMA-1\\~\cite{touvron2023llama}}} & 7B & 33.2 & 0.3 & 31.8 & 0.3 & 58.1 & 50.9 & 45.7 \\
                & 13B & 54.2 & 21.9 & 54.0 & 21.9 & 59.8 & 51.8 & 45.2 \\
                & 30B & 65.4 & 40.2 & 65.4 & 40.2 & 61.3 & 54.5 & 48.6 \\
                & 65B & 64.4 & 38.9 & 64.0 & 37.9 & 63.6 & 55.6 & 50.5 \\
            \midrule
            \multirow{3}{*}{\makecell{LLaMA-2\\~\cite{touvron2023llama2}}} & 7B & 34.6 & 3.2 & 34.6 & 3.2 & 58.6 & 51.9 & 46.9 \\
                    & 13B & 57.7 & 28.5 & 55.4 & 25.5 & 61.9 & 53.9 & 48.4 \\
                    & 70B & 69.8 & 46.7 & 69.3 & 45.5 & 64.7 & 55.3 & 50.1 \\
            \midrule
            \multirow{3}{*}{\makecell{LLaMA-2-Chat\\~\cite{touvron2023llama2}}} & 7B & 60.1 & 33.4 & 60.1 & 33.4 & 57.7 & 50.2 & 45.9 \\
                         & 13B & 65.3 & 37.6 & 65.3 & 37.6 & 58.7 & 50.0 & 46.2 \\
                         & 70B & 74.9 & 55.2 & 74.9 & 55.2 & 61.0 & 54.7 & 48.5 \\
            \midrule
            \multirow{2}{*}{\makecell{WizardLM\\~\cite{xu2023wizardlm}}}
                   & 13B & 67.0 & 42.4 & 67.0 & 42.4 & 60.0 & 50.3 & 47.8 \\
                   & 70B & 74.4 & 55.3 & 74.4 & 55.3 & 60.4 & 53.9 & 49.8 \\
            \midrule
            \multirow{2}{*}{\makecell{Xwin-LM\\~\cite{teamxwin}}} & 7B & 50.9 & 18.3 & 50.9 & 18.3 & 61.1 & 51.4 & 46.9 \\
                 & 13B & 62.2 & 30.2 & 62.2 & 30.1 & 63.6 & 54.1 & 50.1 \\
            \midrule
            \multirow{2}{*}{\makecell{Alpaca\\~\cite{taori2023alpaca}}} & 7B & 54.3 & 21.7 & 54.1 & 21.0 & 57.8 & 51.7 & 47.7 \\
                   & 13B & 56.8 & 29.0 & 56.6 & 28.9 & 60.5 & 50.9 & 46.6 \\
            \midrule
            \multirow{3}{*}{\makecell{Vicuna\\~\cite{chiang2023vicuna}}} & 7B & 60.5 & 39.2 & 60.7 & 39.1 & 57.5 & 49.8 & 45.8 \\
                   & 13B & 67.0 & 51.6 & 67.1 & 51.6 & 61.6 & 51.4 & 47.7 \\
                   & 33B & 69.3 & 46.7 & 69.2 & 46.5 & 60.3 & 50.5 & 46.8 \\
            \bottomrule        
        \end{tabular}
    %}

    \label{tab:CommonSenseQA-0-shot}
\end{table*}

\begin{table*}[t]
    \centering
    \caption{MCQA evaluation on 0-shot RACE (all)~\cite{lai2017race}.}
    \vspace{-3mm}
        \begin{tabular}{ccccccccc}
            \toprule
            Model & Size & Choices & \makecell{Choices\\(Circular)} & Vocab & \makecell{Vocab\\(Circular)} & Alignment & \makecell{Normalized \\Alignment} & PPL \\
             \midrule
             \multirow{2}{*}{\makecell{MPT\\~\cite{team2023introducing, mosaicml2023introducing}}} & 7B & 29.2 & 0.5 & 29.7 & 0.5 & 46.0 & 49.1 & 49.0 \\
                & 30B & 56.8 & 25.9 & 56.6 & 25.7 & 48.5 & 51.7 & 51.2 \\
             \midrule
            \makecell{MPT-Chat\\~\cite{mosaicml2023introducing}} & 30B & 69.2 & 49.7 & 69.4 & 50.0 & 50.4 & 53.5 & 53.2 \\
            \midrule
             \multirow{2}{*}{\makecell{Falcon\\~\cite{almazrouei2023falcon}}} & 7B & 26.5 & 0.0 & 25.3 & 0.0 & 44.1 & 47.3 & 46.6 \\
                & 40B & 66.9 & 43.1 & 65.6 & 42.2 & 49.3 & 52.5 & 51.7 \\
             \midrule
             \multirow{4}{*}{\makecell{LLaMA-1\\~\cite{touvron2023llama}}} & 7B & 36.8 & 3.4 & 33.6 & 2.8 & 46.3 & 50.2 & 48.2 \\
                & 13B & 54.7 & 19.6 & 42.9 & 11.2 & 48.6 & 51.7 & 49.8 \\
                & 30B & 70.2 & 47.2 & 70.0 & 47.1 & 49.6 & 53.0 & 50.8 \\
                & 65B & 75.7 & 54.1 & 71.1 & 50.5 & 52.3 & 55.5 & 53.3 \\
            \midrule
            \multirow{3}{*}{\makecell{LLaMA-2\\~\cite{touvron2023llama2}}} & 7B & 48.2 & 13.0 & 44.4 & 11.5 & 46.8 & 50.2 & 47.9 \\
                    & 13B & 65.9 & 37.3 & 64.0 & 36.0 & 48.1 & 51.9 & 49.5 \\
                    & 70B & 84.8 & 74.2 & 83.3 & 72.3 & 51.9 & 55.3 & 53.1 \\
            \midrule
            \multirow{3}{*}{\makecell{LLaMA-2-Chat\\~\cite{touvron2023llama2}}} & 7B & 64.6 & 37.1 & 64.2 & 36.9 & 49.7 & 52.4 & 50.6 \\
                         & 13B & 72.4 & 49.8 & 72.2 & 49.8 & 53.3 & 56.0 & 54.7 \\
                         & 70B & 83.9 & 71.0 & 83.9 & 71.0 & 56.0 & 57.8 & 56.7 \\
            \midrule
            \multirow{2}{*}{\makecell{WizardLM\\~\cite{xu2023wizardlm}}} 
                   & 13B & 73.5 & 57.2 & 73.5 & 57.1 & 52.7 & 55.9 & 54.3 \\
                   & 70B & 83.2 & 73.1 & 83.2 & 73.0 & 55.6 & 57.5 & 56.3 \\
            \midrule
            \multirow{2}{*}{\makecell{Xwin-LM\\~\cite{teamxwin}}} & 7B & 58.2 & 23.6 & 58.1 & 23.5 & 49.9 & 51.9 & 50.8 \\
                 & 13B & 71.2 & 44.4 & 71.1 & 44.2 & 51.3 & 54.4 & 53.0 \\
            \midrule
            \multirow{2}{*}{\makecell{Alpaca\\~\cite{taori2023alpaca}}} & 7B & 53.7 & 23.8 & 53.5 & 23.3 & 49.4 & 52.0 & 50.9 \\
                   & 13B & 55.0 & 31.2 & 54.7 & 30.6 & 49.8 & 52.9 & 51.5 \\
            \midrule
            \multirow{3}{*}{\makecell{Vicuna\\~\cite{chiang2023vicuna}}} & 7B & 69.3 & 49.9 & 69.3 & 49.9 & 47.6 & 50.7 & 48.8 \\
                   & 13B & 75.7 & 60.7 & 75.7 & 60.7 & 49.4 & 52.8 & 50.8 \\
                   & 33B & 77.8 & 63.2 & 77.7 & 63.0 & 50.6 & 53.4 & 52.5 \\
            \bottomrule        
        \end{tabular}
    %}
    \label{tab:RACE-all-0-shot}
\end{table*}

\begin{table}[!t]
    \centering
    \caption{An example where GPT-4 generates accurate justifications.}
     \vspace{-3mm}
        \begin{tabular}{>{\raggedright\arraybackslash}p{3cm} >{\raggedright\arraybackslash}p{13.5cm}}
            \toprule
            Question & What is the best way to form a new habit? \\
            \midrule
            \makecell[{{p{3cm}}}]{LLM-1's Response\\(Vicuna 13B v1.5)} & \makecell[{{p{13.5cm}}}]
            {1. Start small: Begin with a simple and achievable goal.\\
            2. Make it a routine: Establish a specific time and place for the habit.\\
            3. Track your progress: Keep track of your progress to stay motivated.\\
            4. Reward yourself: Reward yourself for every successful attempt.\\
            5. Stay consistent: Stick to the habit even if you miss a day.\\
            6. Get support: Share your goal with friends or family for support.\\
            7. Reflect on your progress: Reflect on your progress and adjust your approach if needed.}\\
            \midrule
            \makecell[{{p{3cm}}}]{LLM-2's Response\\(Vicuna 7B v1.5)} & \makecell[{{p{13.5cm}}}]
            {1. Start small and make it specific.\\
                2. Make it a daily routine.\\
                3. Get an accountability partner.\\
                4. Reward yourself for success.\\
                5. Be consistent and persistent.\\            
            }\\
            \midrule
            GPT-4's Decision & Tie.\\
            \midrule
            GPT-4's Explanation & \makecell[{{p{13.5cm}}}]
            {Both LLM-1 and LLM-2 provided responses that are accurate and relevant to the user's question about forming a new habit. LLM-1's response is slightly more comprehensive, including seven steps with a bit more detail, such as 'Track your progress' and 'Reflect on your progress.' LLM-2's response is more concise with five steps, but still covers the essential aspects of habit formation. Both responses are clear and easy to understand, adhering to the clarity criterion. There are no ethical or legal standards at risk in this context, so both comply with the compliance criterion. The timeliness criterion is not particularly relevant here, as the strategies for habit formation do not change rapidly over time. Neither response contains misleading or harmful content, and both avoid bias by providing general advice applicable to anyone looking to form a new habit. Given that both responses are satisfactory and there is no significant difference in the quality of the information provided, the evaluation results in a tie.}\\
            \bottomrule        
        \end{tabular}
    %}
    \vspace{-3mm}

    \label{tab:pos-1}
\end{table}

\begin{table}[!t]
    \centering
    \caption{An example where GPT-4 generates accurate justifications.}
    \vspace{-3mm}
        \begin{tabular}{>{\raggedright\arraybackslash}p{3cm} >{\raggedright\arraybackslash}p{13.5cm}}
            \toprule
            Question & What's your favorite part of the video for ``Somebody That I Used to Know" by Walk off the Earth? \\
            \midrule
            \makecell[{{p{3cm}}}]{LLM-1's Response \\ (GPT-4 Turbo)}  & \makecell[{{p{13.5cm}}}]
            {As an AI, I don't have personal preferences or feelings, so I don't have a favorite part. However, the video for ``Somebody That I Used to Know" by Walk off the Earth is widely recognized for its unique and creative approach to music performance. The band members are seen playing a single guitar simultaneously, which is an impressive display of coordination and musical arrangement. This aspect of the video often stands out to viewers and could be considered a highlight by many.}\\
            \midrule
            \makecell[{{p{3cm}}}]{LLM-2's Response \\ (Xwin-LM 13B v0.1)} & \makecell[{{p{13.5cm}}}]
            {The part where they're all playing the same guitar but in different octaves and positions. It's a great example of how creative and innovative they are as musicians.}\\
            \midrule
            GPT-4's Decision & LLM-1 wins.\\
            \midrule
            GPT-4's Explanation & \makecell[{{p{13.5cm}}}]
            {Both LLM-1 and LLM-2 provided relevant responses to the user's question about the music video for 'Somebody That I Used to Know' by Walk off the Earth. LLM-1's response is more effective as it adheres to the criteria of accuracy, relevance, comprehensiveness, and clarity. LLM-1 correctly identifies that as an AI, it does not have personal preferences, and then proceeds to describe a commonly appreciated aspect of the video, which is the band's coordination in playing a single guitar. This response is comprehensive as it provides context to why that part of the video might be considered a highlight. LLM-2's response is relevant and accurate but less comprehensive, as it only mentions the fact that the band members play the same guitar in different octaves and positions without further elaboration or context. Both responses are clear and free from bias, and neither contains harmful content. However, LLM-1's response is more effective overall due to its comprehensive nature.}\\
            \bottomrule        
        \end{tabular}
    %}
     %\vspace{-3mm}
    \label{tab:pos-2}
\end{table}

\begin{table}[!t]
    \centering
    \caption{An example of GPT-4 providing incorrect justifications. We include comments to highlight the inaccuracies.}
     \vspace{-3mm}
        \begin{tabular}{>{\raggedright\arraybackslash}p{3cm} >{\raggedright\arraybackslash}p{13.5cm}}
            \toprule
            Question & Write an email to attendees as a reminder that the event is coming up. \\
            \midrule
            \makecell[{{p{3cm}}}]{LLM-1's Response\\(LLaMA-2 70B Chat)}  & \makecell[{{p{13.5cm}}}]
            {Subject: Reminder: Our Exciting Event is Just Around the Corner!\\
Dear \lbrack Name\rbrack,\\
We hope this email finds you well. With only a few days left until our upcoming event, we wanted to send you a friendly reminder to save the date!\\
Join us on \lbrack Date\rbrack at \lbrack Time\rbrack at \lbrack Location\rbrack for an unforgettable experience filled with \lbrack activities, performances, or keynote speakers\rbrack. We have an amazing lineup of \lbrack insert details\rbrack, and we can't wait for you to be a part of it.\\
Don't miss out on the opportunity to \lbrack insert what attendees will gain from the event, such as networking, learning, or fun\rbrack. Our event is sure to be an exciting and memorable experience for all who attend\\
If you haven't already, please RSVP by \lbrack Date\rbrack to secure your spot. You can do so by visiting our website or by responding to this email.\\
We're looking forward to seeing you there!\\
Best regards,\\
\lbrack Your Name\rbrack\\
\lbrack Your Title\rbrack\\
\lbrack Your Company\rbrack\\
\lbrack Contact Information\rbrack}\\
            \midrule
            \makecell[{{p{3cm}}}]{LLM-2's Response\\(Alpaca 13B)} & \makecell[{{p{13.5cm}}}]
            {\lbrack Empty Reponse\rbrack\\
            }\\
            \midrule
            GPT-4's Decision & Tie (both bad).\\
            \midrule
            GPT-4's Explanation & \makecell[{{p{13.5cm}}}]
            {Both LLM-1 and LLM-2 failed to provide a satisfactory response based on the evaluation criteria. LLM-1 provided a detailed and well-structured email template that is relevant, clear, and comprehensive. However, it lacks specific information such as the actual date, time, location, details about the activities, and the RSVP deadline, which are crucial for the accuracy and timeliness of the communication. LLM-2 did not provide any response at all, which fails all criteria. Therefore, neither response can be deemed effective, resulting in a tie where both responses are unsatisfactory.}\\
            \midrule
            Authors' Comment & The requested email template has been successfully provided by LLM-1. However, it should be noted that GPT-4's critique regarding the absence of specific information needs to be addressed by the user, as LLM-1 is not equipped to do so. Given this limitation, LLM-1 should be acknowledged as the winner.\\
            \bottomrule        
        \end{tabular}
    %}
    \vspace{-3mm}
    \label{tab:neg-1}
\end{table}

\begin{table}[!t]
    \centering
    \caption{An example of GPT-4 providing incorrect justifications. We include comments to highlight the inaccuracies.}
    \vspace{-3mm}
        \begin{tabular}{>{\raggedright\arraybackslash}p{3cm} >{\raggedright\arraybackslash}p{13.5cm}}
            \toprule
            Question & highest number below 100, without using the digit 9 \\
            \midrule
            \makecell[{{p{3cm}}}]{LLM-1's Response\\(Dolly 12B v2)}  & \makecell[{{p{13.5cm}}}]
            {99}\\
            \midrule
            \makecell[{{p{3cm}}}]{LLM-2's Response\\(GPT-3.5 Turbo)} & \makecell[{{p{13.5cm}}}]
            {The highest number below 100 without using the digit 9 is 89.}\\
            \midrule
            GPT-4's Decision & Tie (both bad).\\
            \midrule
            GPT-4's Explanation & \makecell[{{p{13.5cm}}}]
            {LLM-1's response fails to meet the accuracy criterion as it includes the digit 9, which is explicitly against the user's stipulation. LLM-2 correctly identifies 89 as the highest number below 100 that does not include the digit 9, satisfying the accuracy and relevance criteria. Both responses are clear and concise, but LLM-1's response does not comply with the user's request, making LLM-2 the winner. There are no issues with timeliness, harmlessness, or bias in this context.}\\
            \midrule
            Authors' Comment & GPT-4 made the correct decision; however, its justification was erroneous. The statement `LLM-2 accurately determined 89 to be the greatest number under 100 without the digit 9' is incorrect, as 89 does indeed contain the digit 9.\\
            \bottomrule        
        \end{tabular}
    %}

     \vspace{-3mm}
    \label{tab:neg-2}
\end{table}

\section{Examples from RWQ Benchmark}
\label{sec:RWQ}
Our RWQ benchmark comprises 20,772 authentic questions sourced from various platforms including Google Trends, Quora, ShareGPT, LMSYS-Chat-1M, and AlpacaEval. We show three examples for each source in Table~\ref{tab:RWQ-examples}.

\begin{table}[!t]
    \centering
    \caption{Examples from each source in our RWQ benchmark.}
    \vspace{-3mm}
        \begin{tabular}{p{16.5cm}}
            \toprule
            \textit{Google Trends}\\
            Example-1: ``What are some tips for beginners trying to master Wordle?  ''\\
            Example-2: ``Can you name three main characters from the TV show ``Stranger Things''?''\\
            Example-3: ``Did Mac Miller's music style influence the searches trends in hiphop?''\\
             \midrule
            \textit{Quora}\\
            Example-1: ``What are some essential parenting skills?''\\
            Example-2: ``What is the most horrifying noise you have ever heard?''\\
            Example-3: ``What would happen if everyone in the world fell asleep at the same time?''\\
            \midrule
            \textit{AlpacaEval}\\
            Example-1: ``Give me the list of top 100 tech categories.''\\
            Example-2: ``How are carbon fibers used in buildings.''\\
            Example-3: ``Write a 5 verse song in the style of Talking Heads based on the life of a teenager in the 1980s britain.''\\
             \midrule
            \textit{ShareGPT}\\
            Example-1: ``Write me a business plan for my new basketball training company called ProX aka Professional Experience.''\\
            Example-2: ``Give me a physics for problem to test understanding of velocity and gravity.''\\
            Example-3: ``I want you to act as a social media content planner for a new startup company. Please provide ideas for five Instagram posts that showcase the products in different ways. Each post should have a brief caption that highlights the product's unique features or benefits.''\\
             \midrule
            \textit{LMSYS-Chat-1M}\\
            Example-1: ``Can you give me an example of a word that is used almost exclusively in the context of an idiomatic phrase and uncommon otherwise?''\\
            Example-2: ``Conclude what is the meal title from this and what are the total calories (if there is a range, choose higher number). Please print out only meal name and total calories. Act as a nutritionist specialising in educated guessing of total calories for any given meal. Based only on the information you have, make your best educated guess with confidence. If you lack some information for making the conclusion, please guess it. Provide only meal name and total calories and no other text or explanation.''\\
            Example-3: ``Tell me the most common fraud cases in consortium.''\\
            \bottomrule        
        \end{tabular}
    %}
    \label{tab:RWQ-examples}
\end{table}

\end{document}